\documentclass[10pt,twocolumn,letterpaper]{article}

\usepackage[pagenumbers]{wacv}
\usepackage{graphicx}
\usepackage{multirow}
\usepackage{amssymb}
\usepackage{xcolor}
\usepackage[accsupp]{axessibility}

\definecolor{wacvblue}{rgb}{0.21,0.49,0.74}
\usepackage[pagebackref,breaklinks,colorlinks,allcolors=wacvblue]{hyperref}

\title{\texorpdfstring{Stochastic Deep Supervision for Gallery-Consistent Early-Exit Face Recognition}{Stochastic Deep Supervision for Gallery-Consistent Early-Exit Face Recognition}}

\author{Guray Ozgur$^{1,2}$ \quad Tahar Chettaoui$^{1,2}$ \quad Eduarda Caldeira$^{1,2}$ \quad Naser Damer$^{1,2}$ \quad Fadi Boutros$^{2}$\\
$^{1}$TU Darmstadt, Darmstadt, Germany\\
$^{2}$Fraunhofer Institute for Computer Graphics Research IGD, Darmstadt, Germany\\
{\tt\small guray.ozgur@igd.fraunhofer.de}
}

\begin{document}
\maketitle

\begin{abstract}
Vision Transformers (ViTs) perform strongly in face recognition (FR), yet their fixed-depth inference limits deployment in resource-constrained settings. We propose SD-ViT, a training framework that gives a usable exit at every block depth by making every depth produce discriminative embeddings, for budget-selected fixed-depth exits. Standard ViTs supervise intermediate layers only through backpropagation, so their intermediate representations are not directly optimized to form usable identity embeddings. SD-ViT adds a stochastic auxiliary signal: at each iteration a randomly selected intermediate block is passed through the same embedding head as the final layer and trained with a weighted margin-penalty softmax loss. The objective is thus stochastically sampled deep supervision through a \emph{shared} classifier. Because both loss terms share that head and classifier, intermediate tokens are trained to map into the same embedding space as final-layer tokens. This matters for FR, where matching is by similarity: a probe processed at an early exit can be compared directly against full-depth gallery embeddings, with no gallery recomputation or adaptation layers. On established benchmarks, SD-ViT stays within $0.4$ IJB-C points of the same backbone trained conventionally at full depth, while its early-exit probes recover over $96\%$ of their own full-depth performance at $50\%$ of the computational cost without modifying the reference gallery. A single verification threshold calibrated at full depth remains usable from quarter depth onward. Code: \url{https://github.com/gurayozgur/SD-ViT}.
\end{abstract}

\begin{figure*}[t]
\centering
\includegraphics[width=\linewidth]{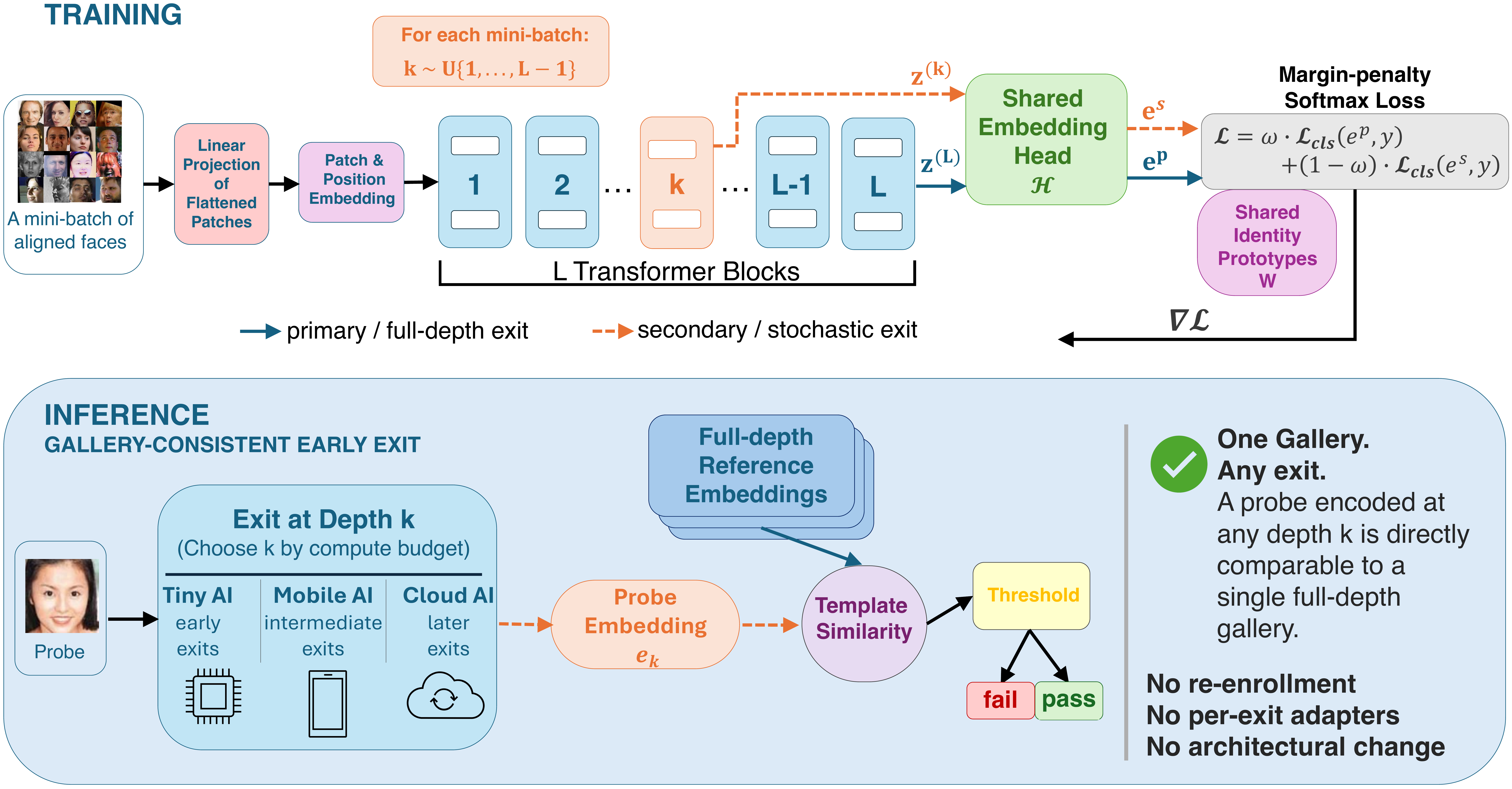}
\caption{Overview of SD-ViT. A face image is tokenized and processed by $L$ blocks. The full-depth tokens $z^{(L)}$ form the primary exit. At each iteration a secondary exit is read out at a uniformly sampled block $z^{(k)}$, $k \sim \mathcal{U}\{1,\dots,L-1\}$. Both are decoded by the \emph{same} head $\mathcal{H}$ and scored against the \emph{same} prototypes $W$ under a margin-penalty loss, combined with weight $\omega$ (Eq.~\ref{eq:loss}). Sharing $\mathcal{H}$ and $W$ across depths places all embeddings in one space, so at inference a probe may exit at any depth $k$ (Eq.~\ref{eq:infer}) and match a gallery enrolled once at full depth.}
\label{fig:overview}
\end{figure*}

\section{Introduction}
\label{sec:intro}
FR accuracy is driven by deep networks \cite{ResNet} trained with margin-based softmax objectives \cite{deng2019arcface,DBLP:conf/cvpr/WangWZJGZL018,DBLP:conf/cvpr/HuangWT0SLLH20}, and increasingly by ViTs \cite{vit_dosovitskiy}, which surpass convolutional backbones on FR benchmarks \cite{TransFace,adaface,kim2024kprpe}. A ViT processes every input through a fixed stack of $L$ blocks at the same cost regardless of input difficulty or device budget, which is at odds with deployment: one model must run from edge processors to datacenter accelerators under budgets differing by orders of magnitude \cite{DBLP:conf/iccvw/DengGZDLS19}. Compact architectures \cite{MFN,MixFaceNets,DBLP:conf/iccvw/YanZXZWS19}, compression \cite{DBLP:conf/icpr/BoutrosDK22} and distillation \cite{KD,ReFO,EKD} each yield another \emph{static} model from a separate run, so covering several budgets means training, storing and switching whole networks. Early-exit networks are the flexible alternative: one model with intermediate classifiers, so inference stops shallow for easy inputs or tight budgets \cite{branchynet,msdnet}. They were built for closed-set classification, where each exit has its own classifier over a fixed label set and the exits need not share a representation \cite{branchynet,msdnet,kaya2019sdn}. FR does not fit that template: it is open-set, matching a probe to a gallery of enrolled embeddings by cosine similarity, so independent per-exit heads produce incompatible spaces and a shallow embedding cannot match a full-depth gallery. The closed-set and distillation analogues, per-exit heads that would require re-embedding the gallery at every depth \cite{bakhtiarnia2021multiexit} and learned adaptation layers between depths \cite{FitNet}, are unattractive here. Galleries can be massive, distributed or privacy-constrained, and adaptation layers break the single-embedding-space assumption enrollment pipelines rely on. Early-exit FR therefore requires that \emph{every} depth emit embeddings in one shared space, compatible with a gallery enrolled once at full depth.

We address this with \textbf{S}tochastic \textbf{D}eep-supervision for ViTs (SD-ViT), a \emph{shared-classifier} form of deep supervision (Fig.~\ref{fig:overview}). At each iteration, besides the full-depth output, SD-ViT reads out one auxiliary embedding at a \emph{uniformly sampled} intermediate block, decodes it with the \emph{same} head, and scores it against the \emph{same} prototypes under the same margin-penalty loss. The shared head maps tokens from every depth into one embedding space, and the shared prototypes give each identity a single center in it. Deep supervision is long established \cite{DSN,GoogLeNet}. What distinguishes SD-ViT is not that intermediate depths are supervised but \emph{through what}: the final exit's own head and classifier instead of exit-specific ones, which makes the embeddings interchangeable across depth. Our controls show that the sampled exit must be trained by the margin loss itself, not by feature matching (Sec.~\ref{sec:ablation}). At inference a probe may exit at any depth and match a full-depth gallery by plain cosine similarity. Sampling a different block each step spreads supervision uniformly over depth, so every block depth becomes a usable exit after one training run. Sampling is a single-sample estimator of supervising all exits, and Sec.~\ref{sec:ablation} finds it on par with deterministic all-exit supervision at one seventh of the auxiliary cost. We contribute: (i)~the identification of \emph{gallery consistency}, one shared embedding space across depths, as the central obstacle to early-exit FR, (ii)~SD-ViT, in which a uniformly sampled auxiliary exit shares the final layer's head and prototypes, making every depth gallery-compatible with no teacher and no adaptation layers, and (iii)~evidence that one training run yields a usable exit at every block depth, with full-depth accuracy within $0.4$ IJB-C points of conventional training, no gallery recomputation, and a single verification threshold usable from quarter depth onward.

\section{Related Work}
\label{sec:related}

\textbf{Identity objectives for FR.}
Margin-penalty softmax losses \cite{deng2019arcface,DBLP:conf/cvpr/WangWZJGZL018,adaface,ElasticFace} and auxiliary objectives \cite{centerloss, gitloss} separate identities and compact each around its center, but only for the final-layer embedding. SD-ViT applies such supervision at a sampled intermediate depth as well, against the prototypes that define the final space.

\textbf{Deep supervision and knowledge distillation.}
Deeply-supervised nets \cite{DSN} and GoogLeNet \cite{GoogLeNet} add auxiliary losses at intermediate layers to aid optimisation and discard the auxiliary heads after training. Knowledge distillation \cite{KD} transfers from teacher to student via soft-label or feature-matching objectives, and born-again networks \cite{born_again} and deep mutual learning \cite{dml} remove the fixed teacher. For FR it has used triplet \cite{TripletDistillation}, rank-based \cite{EKD}, angular \cite{MarginKD} and reverse-feature \cite{ReFO} objectives, and AdaDistill \cite{boutros2024adadistill} injects teacher-derived class centers into the student margin loss. BYOT \cite{byot} combines the two: it attaches auxiliary classifiers to intermediate CNN sections and distills from the deepest into the shallower ones. SD-ViT differs from both lines. Its auxiliary exit is not discarded after training but deployed as an early exit, and it is decoded by the final exit's own head and scored against the same prototypes instead of a separate classifier. It uses no teacher and no adaptation layers, and couples the exits solely through this shared head and prototypes.

\textbf{Early-exit and budget-selected inference.}
Early-exit networks add intermediate classifiers so that easy inputs terminate early \cite{branchynet,msdnet,kaya2019sdn,zhou2020pabee}. LayerSkip \cite{layerskip} is the closest precedent for our parameter sharing: it trains a language model with layer dropout \cite{stochasticdepth} and an early-exit loss through a \emph{shared} output head, so intermediate layers become decodable without per-layer heads. Among early-exit ViTs, LGViT \cite{lgvit} uses heterogeneous local-global heads and EnViT \cite{envit} exit-aware dropout, both trained by self-distillation. All target closed-set classification, where the label set ties the exits together and nothing forces their \emph{representations} to be comparable. Open-set FR adds that requirement. Token pruning and merging \cite{rao2021dynamicvit,bolya2023tome} reduce the cost of each block, not the number of blocks, so they are orthogonal to SD-ViT and could be combined with it. Closest to our setting, ViT-FREE \cite{vitfree} performs \emph{training-free} early exiting in pretrained ViT-FR models, exploiting constant token dimensionality with a synthetic-data adaptation step, and EX-FIQA \cite{exfiqa} fuses intermediate representations for quality assessment. SD-ViT shares the observation that a ViT keeps one dimensionality at every depth, but instead of reading exits from a \emph{fixed} backbone post-hoc it optimizes \emph{every} block to produce gallery-consistent embeddings. We compare against both of its settings (Sec.~\ref{sec:main_results}). Training-free ViT-FREE is precisely our Baseline read-out. ViT-FREE$_{FT}$ trains an exit-specific projection and prototype set per depth. We report it under self-evaluation, the setting it was designed for, and under our gallery-fixed setting.

\textbf{ViTs for FR.}
TransFace \cite{TransFace} addresses ViT overfitting with patch-level augmentation and hard-sample mining, Part-fViT \cite{sun2022partfvit} adds landmark-driven part tokens, and KP-RPE \cite{kim2024kprpe} replaces learnable positional encodings with keypoint-relative ones. All treat the ViT as a fixed-depth backbone for a single full-depth pass, addressing neither multi-exit inference nor gallery-consistent retrieval. SD-ViT adds a stochastic shared-classifier deep-supervision objective that makes every block produce gallery-compatible embeddings, with no change to the inference architecture.

\begin{table}[t]
\centering
\caption{Baseline vs.\ SD-ViT at quarter-, half-, and full-depth exits, on WebFace4M (WF4M) and WebFace12M (WF12M) \cite{webface4m}. \emph{Baseline} is the identical backbone trained identically except that supervision is applied at the final exit only ($\omega{=}1$, no auxiliary term). Reading its intermediate blocks through the shared head is the training-free ViT-FREE \cite{vitfree} protocol. \emph{ViT-FREE$_{FT}$} trains one projection and one prototype set per exit on that frozen Baseline. \emph{Truncated} are ViTs of 3 and 6 blocks trained from scratch, i.e.\ dedicated single-budget networks, so each of their rows is that network's only exit. HQ\,(5) averages the five high-quality sets, TinyFace reports Rank-1, and IJB-C TAR@FAR$=10^{-4}$ (Sec.~\ref{sec:setup}). GFLOPs is the exit cost. Bold: the better of Baseline and SD-ViT.}
\label{tab:main_results}
\scriptsize
\setlength{\tabcolsep}{3pt}
\resizebox{\linewidth}{!}{%
\begin{tabular}{lll|ccccc}
\hline
 & & & Exit & GFLOPs & HQ\,(5) & TinyFace & IJB-C \\
Backbone & Data & Method & $k/L$ & & avg & R-1 & $10^{-4}$ \\
\hline
\multirow{11}{*}{ViT-S/12} & \multirow{11}{*}{WF4M}
 & \multirow{3}{*}{Baseline}
   & 3/12  & 4.70  & 63.68 & 40.61 & 11.05 \\
 & & & 6/12  & 9.26  & 77.90 & 56.01 & 42.22 \\
 & & & 12/12 & 18.37 & \textbf{97.43} & \textbf{74.25} & \textbf{96.89} \\
\cline{3-8}
 & & \multirow{3}{*}{SD-ViT}
   & 3/12  & 4.70  & \textbf{93.68} & \textbf{64.62} & \textbf{87.01} \\
 & & & 6/12  & 9.26  & \textbf{96.42} & \textbf{71.22} & \textbf{95.11} \\
 & & & 12/12 & 18.37 & 97.29 & 73.85 & 96.51 \\
\cline{3-8}
 & & \multirow{3}{*}{ViT-FREE$_{FT}$}
   & 3/12  & 4.70  & 88.91 & 57.56 & 71.36 \\
 & & & 6/12  & 9.26  & 94.68 & 67.09 & 89.74 \\
 & & & 12/12 & 18.37 & 97.39 & 73.39 & 96.74 \\
\cline{3-8}
 & & \multirow{2}{*}{truncated}
   & 3/3   & 4.70  & 95.86 & 69.50 & 93.78 \\
 & & & 6/6   & 9.26  & 96.96 & 71.83 & 95.94 \\
\hline
\multirow{6}{*}{ViT-B/24} & \multirow{6}{*}{WF4M}
 & \multirow{3}{*}{Baseline}
   & 6/24  & 6.79  & 69.35 & 49.73 & 21.63 \\
 & & & 12/24 & 13.44 & 88.35 & 65.45 & 71.97 \\
 & & & 24/24 & 26.74 & \textbf{97.45} & 73.77 & \textbf{97.02} \\
\cline{3-8}
 & & \multirow{3}{*}{SD-ViT}
   & 6/24  & 6.79  & \textbf{95.41} & \textbf{68.51} & \textbf{91.83} \\
 & & & 12/24 & 13.44 & \textbf{97.05} & \textbf{73.50} & \textbf{96.32} \\
 & & & 24/24 & 26.74 & 97.35 & \textbf{74.57} & 96.69 \\
\hline
\multirow{6}{*}{ViT-B/24} & \multirow{6}{*}{WF12M}
 & \multirow{3}{*}{Baseline}
   & 6/24  & 6.79  & 64.46 & 44.21 & 12.10 \\
 & & & 12/24 & 13.44 & 82.69 & 63.17 & 55.93 \\
 & & & 24/24 & 26.74 & \textbf{97.75} & \textbf{75.78} & \textbf{97.69} \\
\cline{3-8}
 & & \multirow{3}{*}{SD-ViT}
   & 6/24  & 6.79  & \textbf{96.38} & \textbf{70.04} & \textbf{95.03} \\
 & & & 12/24 & 13.44 & \textbf{97.42} & \textbf{74.41} & \textbf{97.06} \\
 & & & 24/24 & 26.74 & 97.67 & 74.60 & 97.56 \\
\hline
\end{tabular}}
\end{table}

\section{Methodology}
\label{sec:method}
SD-ViT trains a single ViT-FR model whose \emph{every} depth produces a discriminative identity embedding. We review the backbone and margin-penalty loss (Sec.~\ref{sec:prelim}), present SD-ViT (Sec.~\ref{sec:sd}), and describe gallery-consistent early-exit inference (Sec.~\ref{sec:inference}), Fig.~\ref{fig:overview} gives an overview.

\subsection{Preliminaries}
\label{sec:prelim}

\textbf{ViT-FR backbone.}
A ViT-FR backbone \cite{vit_dosovitskiy,TransFace} splits $x$ into $N$ patches, projects each to a $d$ token, adds positional embeddings, and applies $L$ pre-normalized blocks $z^{(l)}=\mathcal{B}_l(z^{(l-1)})$. A token sequence $u$ becomes an embedding through the head $\mathcal{H}$,
\begin{equation}
\label{eq:head}
e = \mathcal{H}(u) = \phi\!\left(\mathrm{vec}\!\left(\mathrm{LN}(u)\right)\right) \in \mathbb{R}^{D},
\end{equation}
where $\mathrm{LN}$ is a network-wide layer normalization, $\mathrm{vec}$ flattens the tokens, and $\phi$ is an $Nd\!\to\!d\!\to\!D$ projection with batch normalization. We write $\hat{e}=e/\|e\|$. The standard setup produces only $e_L = \mathcal{H}(z^{(L)})$.

\textbf{Margin-penalty softmax loss.}
The backbone is trained as a closed-set classifier over $C$ identities \cite{deng2019arcface,DBLP:conf/cvpr/WangWZJGZL018,ElasticFace}. With class prototypes $W=[w_1,\dots,w_C]$, $\ell_2$-normalized columns $\hat{w}_j$ and $\cos\theta_j=\hat{w}_j^{\top}\hat{e}$, the margin-softmax loss is
\begin{equation}
\label{eq:cls}
\mathcal{L}_{\mathrm{cls}}(e, y) = -\log \frac{\exp\!\big(s\,\psi(\theta_{y})\big)}{\exp\!\big(s\,\psi(\theta_{y})\big) + \sum\limits_{j \neq y}\exp\!\big(s\,\cos\theta_{j}\big)},
\end{equation}
with $y$ the ground-truth identity, $s$ a scale and $\psi$ a margin on the target logit (e.g.\ ArcFace \cite{deng2019arcface} or CosFace \cite{DBLP:conf/cvpr/WangWZJGZL018}). SD-ViT applies to any such $\mathcal{L}_{\mathrm{cls}}$. We use AdaFace \cite{adaface} and show in Sec.~\ref{sec:ablation} that ArcFace gives the same pattern.

\textbf{Why intermediate depths are not embeddings.}
Only $e_L$ is supervised, intermediate $z^{(l)}$ receive gradients through backpropagation alone and are never required to be separable or aligned to $W$, so reading them out does not yield an embedding matchable against full-depth references.

\subsection{\texorpdfstring{Stochastic Shared-Classifier Deep Supervision}{Stochastic Shared-Classifier Deep Supervision}}
\label{sec:sd}

\textbf{Two exits sharing one embedding space.}
At every training iteration we keep the full-depth (primary) embedding and additionally read out a secondary embedding at a randomly chosen block,
\begin{equation}
\label{eq:exits}
e^{\mathrm{p}} = \mathcal{H}\!\left(z^{(L)}\right), \qquad
e^{\mathrm{s}} = \mathcal{H}\!\left(z^{(k)}\right), \quad k \sim \mathcal{U}\{1,\dots,L-1\},
\end{equation}
with $\mathcal{U}$ uniform over the first $L-1$ blocks. Both exits are decoded by the \emph{shared} head $\mathcal{H}$, including its $\mathrm{LN}$, and scored against the \emph{shared} prototypes $W$. Exposed to tokens from many depths, one $\mathrm{LN}$ and one $\phi$ must serve every block. Because a ViT keeps token dimensionality $N\times d$ at every block, unlike the changing width of CNN stages, the same head applies unchanged at any depth, with \emph{no} dimensionality reduction and \emph{no} per-exit adaptation layers. SD-ViT targets backbones of constant token width, and Sec.~\ref{sec:ablation} verifies the objective on a second such backbone, KP-RPE \cite{kim2024kprpe}.

\textbf{Weighted dual objective.}
Both exits are supervised by the same margin-penalty loss, combined with a fixed weight $\omega$:
\begin{equation}
\label{eq:loss}
\mathcal{L} = \omega\,\mathcal{L}_{\mathrm{cls}}\!\left(e^{\mathrm{p}}, y\right) + (1-\omega)\,\mathcal{L}_{\mathrm{cls}}\!\left(e^{\mathrm{s}}, y\right).
\end{equation}
The primary exit gets the larger weight, keeping full-depth accuracy the principal objective, while the secondary term supervises the sampled depth against the same $W$.

\textbf{Why the exits become comparable.}
Eq.~\ref{eq:loss} never asks $e^{\mathrm{s}}$ to match $e^{\mathrm{p}}$: there is no teacher, no detached target and no divergence term between the exits. Both terms are identity-label losses that update the backbone, $\mathcal{H}$ and $W$, and $\omega$ only balances them. Whatever aligns the two exits must therefore come from what the two terms share, and we see two candidate effects.

The first is the shared head: $\mathcal{H}$ is one map, one $\mathrm{LN}$, one projection $\phi$ and one set of batch-normalization statistics, that must turn the final tokens and the tokens of every sampled block into embeddings the margin loss can separate. At test time $\phi$ is affine, so $e_L - e_k$ is the linear part of $\phi$ applied to the difference of the two normalized token states. In a pre-norm ViT those states are two snapshots of one residual stream, $z^{(L)} = z^{(k)} + \sum_{l>k}\Delta_l$. Making both discriminative through the same $\phi$ favors later blocks that refine the identity information $\phi$ already reads at depth $k$, and does not re-encode it, which would keep $e_k$ close to $e_L$. Per-exit heads remove this pressure, since each can read identity from a different subspace.

The second is the shared prototypes. $\mathcal{L}_{\mathrm{cls}}$ depends on $e$ \emph{only through} the cosines $\{\hat{w}_j^{\top}\hat{e}\}_{j=1}^{C}$ and is minimized by driving $\hat{e}$ toward $\hat{w}_y$, so with one $W$ the embeddings of identity $y$ at every supervised depth are pulled toward one center. With the head shared, a separate prototype set for the intermediate exits costs about a point (Sec.~\ref{sec:ablation}). Both come from a \emph{single} forward pass, since $z^{(k)}$ is produced en route to $z^{(L)}$, so the auxiliary path adds one head-and-classifier evaluation.

\textbf{Shared statistics and prototype sampling.}
Three details follow from sharing $\mathcal{H}$ and $W$. (i)~The batch-normalization layers of $\phi$ are traversed twice per step. Each branch is normalized by its own batch statistics and both updates enter the same running estimates, so the deployed statistics mix the final and the sampled depths and one set serves every exit at test time. (ii)~With PartialFC \cite{DBLP:conf/iccvw/AnZGXZFWQZZF21} both terms of Eq.~\ref{eq:loss} see the \emph{same} positive prototypes, since the sampled subset always contains the batch labels. Negatives are redrawn per term, and both terms update the same $W$. (iii)~AdaFace's running feature-norm statistics, which set its adaptive margin, are updated by both terms.

\subsection{Early-Exit Inference}
\label{sec:inference}

At test time the model terminates at any depth $k$ and reads out an embedding through the same head,
\begin{equation}
\label{eq:infer}
e_k = \mathcal{H}\!\left(z^{(k)}\right), \qquad k \in \{1,\dots,L\},
\end{equation}
where $k{=}L$ recovers the standard full-depth model.

\textbf{Gallery consistency.}
SD-ViT's defining property for FR is that early-exit probes require \emph{no} change to the gallery. Since $\mathcal{H}$ and $W$ are shared, $\hat{e}_k$ lies in the same space as a full-depth gallery embedding, so cosine similarity stays meaningful between a shallow probe and deep references: no per-exit projection, no re-enrollment. The property holds within one model, with probe and gallery from the same network for the constant-width ViTs. An operational system also needs its threshold to survive the change of depth. A ROC-based metric does not establish this, since TAR at a target FAR may pick a different threshold per exit. Sec.~\ref{sec:gallery} tests the stronger property: one threshold, calibrated on full-depth development scores, applied unchanged at every exit.

\textbf{Computational scaling.}
Exiting at block $k$ runs $k$ of $L$ blocks, cutting backbone computation to about $k/L$, while the head runs on all $N$ tokens at a fixed cost. Choosing $k$ trades accuracy for latency within one model, with no retraining. Throughout, $k$ is selected by a compute budget, not per input: we evaluate budget-selected fixed-depth exits, not an input-adaptive policy. A per-image policy is left to future work.

\begin{figure*}[t]
\centering
\includegraphics[width=\linewidth]{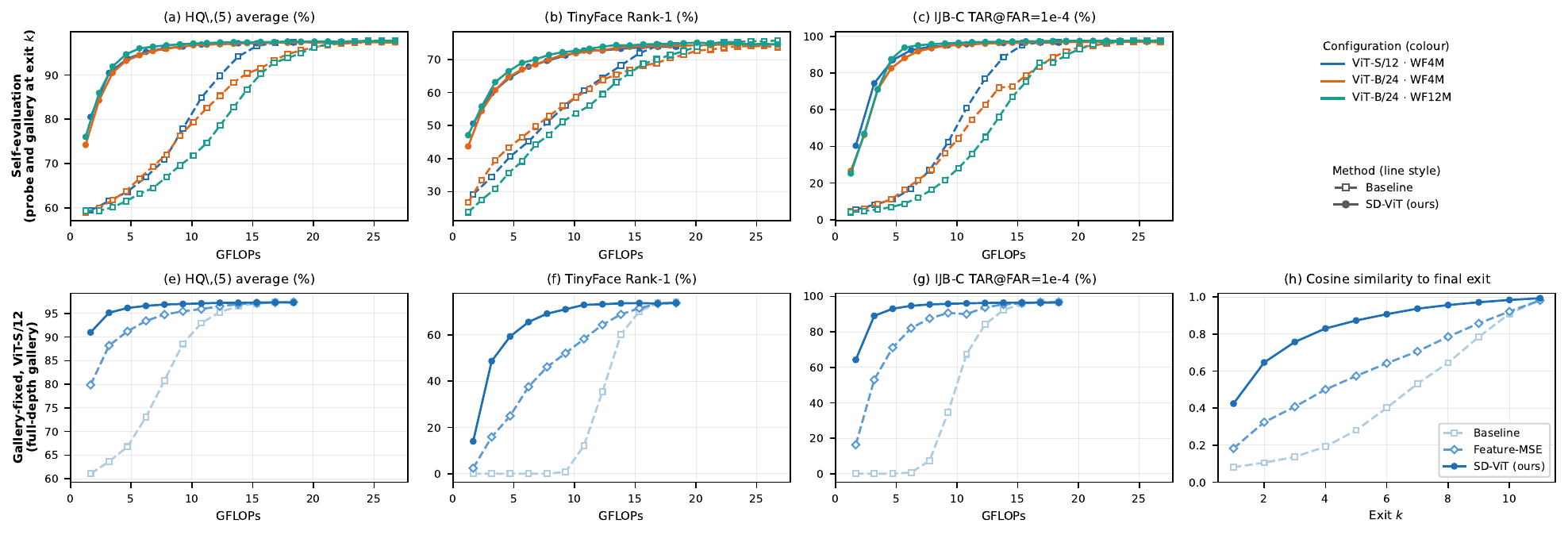}
\caption{Per-exit accuracy under both protocols. \textbf{Top (a-c)}: self-evaluation, with probe and gallery embedded at the same exit, for three configurations. Color denotes the configuration and line style the method (see the legend panel). Baseline early exits collapse while SD-ViT degrades far less, on every backbone and both training sets. \textbf{Bottom (e-g)}: the gallery-fixed protocol of Sec.~\ref{sec:gallery}, an exit-$k$ probe against a gallery enrolled once at the final block, for ViT-S/12. Here the Baseline is near zero on the harder benchmarks below half depth, and Feature-MSE lies between it and SD-ViT. \textbf{(h)}: mean cosine similarity between each exit and the final-block embedding, averaged over the test sets. SD-ViT aligns intermediate embeddings more than the explicit feature-matching variant.}
\label{fig:main_results}
\end{figure*}

\begin{figure*}[t]
\centering
\includegraphics[width=\linewidth]{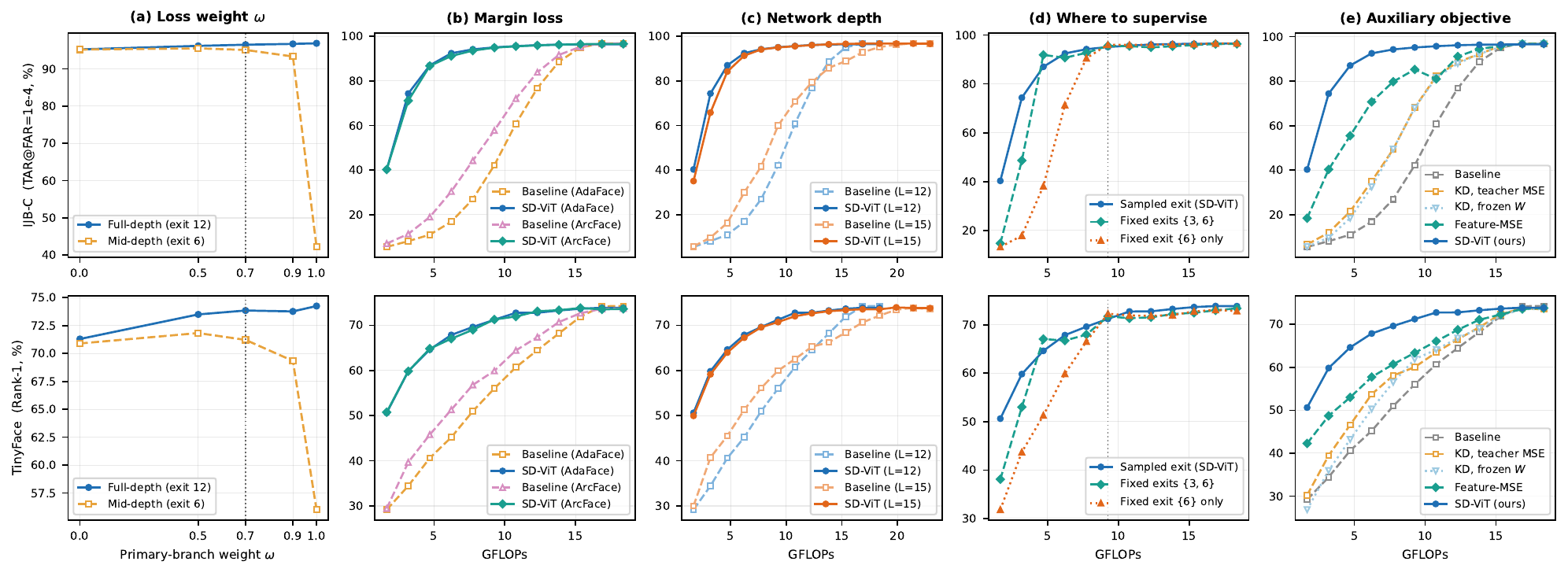}
\caption{Ablations on ViT-S/12, WebFace4M (IJB-C TAR@FAR$=10^{-4}$ top, TinyFace Rank-1 bottom). \textbf{(a)} \textbf{Weight $\omega$:} full depth (solid) is nearly constant for $\omega\ge0.5$, mid-depth (dashed) stable for $\omega\le0.9$ and collapsing at $\omega{=}1$, dotted marks $\omega{=}0.7$. \textbf{(b)} \textbf{Loss:} comparable early-exit accuracy under AdaFace and ArcFace. \textbf{(c)} \textbf{Depth:} the gains hold at $L{=}15$. \textbf{(d)} \textbf{Where to supervise:} sampling degrades smoothly toward exit 1, while fixed exits collapse below the chosen depths. \textbf{(e)} \textbf{Auxiliary objective:} teacher KD and feature-MSE raise early exits above the baseline yet collapse at shallow depths, while SD-ViT stays far above them at early exits.}
\label{fig:ablations}
\end{figure*}

\section{Experiments}
\label{sec:exp}

\subsection{Experimental Setup}
\label{sec:setup}

\textbf{Training data.}
We train on WebFace4M \cite{webface4m} ($4.2$M images, $206$K ids) and, for scaling, ViT-B on WebFace12M \cite{webface4m} ($12.7$M, $618$K).

\textbf{Architectures.}
ViT-S/12, ViT-S/15 and ViT-B/24 on $112\times112$ images, at $18.37$/$22.93$/$26.74$ full-depth GFLOPs. GFLOPs follow \cite{DBLP:conf/kdd/RasleyRRH20}, with $8\times8$ patches and $d{=}512$ throughout. The patch embedding and head add the same $0.14$ GFLOPs at every exit on both backbones, and only the per-block cost differs ($1.52$ ViT-S, $1.11$ ViT-B).

\textbf{Loss and training.}
We adopt AdaFace \cite{adaface} ($m{=}0.4$, $h{=}0.333$) with the augmentations of \cite{adaface}: 40 epochs, batch $1024$, AdamW \cite{adamw} (lr $10^{-4}$, weight decay $0.05$), a cosine schedule \cite{loshchilov2016sgdr,rw2019timm} with 3-epoch warmup, and PartialFC \cite{DBLP:conf/iccvw/AnZGXZFWQZZF21,DBLP:conf/cvpr/AnDGFZ0L22} at sampling rate $0.6$. Unless stated otherwise $\omega{=}0.7$ (Sec.~\ref{sec:sd}). It was chosen on the reporting benchmarks (Tab.~\ref{tab:ablation}). An ArcFace \cite{deng2019arcface} variant tests a second margin loss. The auxiliary exit adds $3$-$6\%$ training time ($65.5\!\to\!68.6$ A100 GPU-h for ViT-S/12).

\textbf{Baseline and reference models.}
\emph{Baseline} is the \emph{same} backbone, data, augmentation, optimizer, schedule, PartialFC setting, margin loss, batch and seed as SD-ViT. The only difference is Eq.~\ref{eq:loss} with $\omega{=}1$ and no auxiliary term. Its architecture and per-exit cost are identical, so the differences below isolate the objective. Reading the Baseline's intermediate blocks through the shared head is exactly the training-free protocol of ViT-FREE \cite{vitfree}, so its rows are simultaneously that comparison. Tab.~\ref{tab:main_results} adds ViT-FREE$_{FT}$ and truncated ViTs at matched GFLOPs. ViT-FREE$_{FT}$ follows \cite{vitfree}, fine-tuning one projection head per exit with its own classifier on the frozen backbone. For a like-for-like comparison, that backbone is our Baseline and the heads are trained with its data and loss (WebFace4M, AdaFace). It is evaluated under self-evaluation in Tab.~\ref{tab:main_results} and \ref{tab:mechanism}, and gallery-fixed in Tab.~\ref{tab:gallery}.

\textbf{Benchmarks and protocol.}
We report the five high-quality sets \cite{LFWTech,agedb,cfp-fp,CPLFWTech,CALFW} by their average, TinyFace \cite{tinyface} Rank-1 and IJB-C \cite{ijbc} TAR@FAR$=10^{-4}$. To evaluate exit $k$ we run the first $k$ blocks, apply the shared head, $\ell_2$-normalize and match by cosine similarity, either between exit-$k$ embeddings (default) or from an exit-$k$ probe to a full-depth reference (gallery-fixed). In the latter the enrolled side is always full depth and the querying side exit $k$. For IJB-C every image of an enrollment template, the first template of each pair, is embedded at full depth and the rest are probes.

\subsection{Early-Exit Recognition across Depths}
\label{sec:main_results}

Tab.~\ref{tab:main_results} and Fig.~\ref{fig:main_results} compare final-exit-only training against SD-ViT. Because a standard ViT never decodes intermediate blocks into embeddings, its early exits collapse: the ViT-S/12 baseline reaches $42.2\%$ IJB-C and $56.0\%$ TinyFace at half depth, and $11.1\%$ and $40.6\%$ at quarter depth. SD-ViT reaches $95.1\%$/$71.2\%$ and $87.0\%$/$64.6\%$ at those same budgets ($9.26$ and $4.70$ against $18.37$ GFLOPs). The pattern repeats on ViT-B/24 and at scale: on WebFace12M the baseline gives $12.1\%$ IJB-C at quarter depth where SD-ViT gives $95.0\%$. Full depth costs a little. IJB-C is $0.38$, $0.33$ and $0.13$ points below the Baseline in the three configurations, and TinyFace on WebFace12M drops $1.18$ points. There is \emph{no} inference overhead, since the two architectures are identical.

\textbf{Against dedicated single-budget models.} Tab.~\ref{tab:main_results} also reports ViTs of $3$ and $6$ blocks trained from scratch at the GFLOPs of SD-ViT's quarter- and half-depth exits. They are better at their own budget: $93.78$ vs.\ $87.01$ IJB-C at $4.70$ GFLOPs and $95.94$ vs.\ $95.11$ at $9.26$. The trade is not accuracy at one budget but coverage: $N$ budgets otherwise need $N$ networks. In FR each has its own embedding space, so a gallery enrolled with one cannot be queried by another. SD-ViT covers every budget in one network against one gallery, giving up $0.83$ points at half depth to do so.

\subsection{Ablation Studies}
\label{sec:ablation}

Ablations use ViT-S/12 on WebFace4M with AdaFace unless stated (Fig.~\ref{fig:ablations}, Tab.~\ref{tab:ablation}).

\begin{table}[t]
\centering
\caption{Primary-loss weight $\omega$ and margin-penalty loss on ViT-S/12, WebFace4M, at mid-depth (exit 6, $9.26$ GFLOPs) and full depth (exit 12, $18.37$ GFLOPs). $\omega=1.0$ is the no-auxiliary baseline, $\omega=0.0$ supervises only the sampled exit. Metrics as in Sec.~\ref{sec:setup}.}
\label{tab:ablation}
\scriptsize
\setlength{\tabcolsep}{2.6pt}
\begin{tabular}{ll|ccc|ccc}
\hline
 & & \multicolumn{3}{c|}{Mid-depth (exit 6)} & \multicolumn{3}{c}{Full depth (exit 12)} \\
Loss & $\omega$ & HQ & TF & IJB-C & HQ & TF & IJB-C \\
\hline
AdaFace & 0.0 & 96.51 & 70.90 & 95.21 & 96.61 & 71.30 & 95.25 \\
AdaFace & 0.5 & 96.70 & 71.83 & 95.52 & 97.06 & 73.50 & 96.20 \\
AdaFace & \textbf{0.7} & 96.42 & 71.22 & 95.11 & 97.29 & 73.85 & 96.51 \\
AdaFace & 0.9 & 95.53 & 69.34 & 93.35 & 97.32 & 73.77 & 96.73 \\
AdaFace & 1.0 (base) & 77.90 & 56.01 & 42.22 & 97.43 & 74.25 & 96.89 \\
\hline
ArcFace & 0.7 & 96.35 & 71.27 & 94.97 & 97.20 & 73.61 & 96.65 \\
ArcFace & 1.0 (base) & 83.41 & 59.92 & 57.82 & 97.41 & 73.71 & 96.93 \\
\hline
\end{tabular}
\end{table}

\textbf{Effect of the primary-loss weight $\omega$.}
Full depth varies at most $0.53$ points on IJB-C over $\omega\in\{0.5,0.7,0.9\}$, while early exits degrade sharply as $\omega\!\to\!1$: half-depth IJB-C falls $95.52\!\to\!95.11\!\to\!93.35\!\to\!42.22\%$. At $\omega{=}0$ early exits stay strong but full depth drops to $95.25\%$ against $96.51\%$, so a dominant primary term is needed. We use $\omega{=}0.7$.

\textbf{Margin loss.}
The gains are not specific to AdaFace. The half-depth IJB-C ArcFace baseline ($57.82\%$) is higher than AdaFace's ($42.22\%$), yet SD-ViT brings both to comparable levels, $94.97\%$ and $95.11\%$, gains of $+37.2$ and $+52.9$ points, with full depth within $0.4$ points of the baseline in both.

\textbf{Network depth.}
On a ViT-S/15 the pattern holds: at half depth the baseline reaches $70.81\%$ IJB-C and $62.53\%$ TinyFace against SD-ViT's $95.59\%$ and $72.0\%$, with full depth preserved ($96.70$ vs.\ $96.83\%$).

\textbf{A second ViT backbone.} To check that the objective is not specific to a plain ViT, we repeat it on KP-RPE \cite{kim2024kprpe}, which replaces learnable positional encodings with keypoint-relative ones and is itself a stronger full-depth model ($97.03$ vs.\ $96.89$ IJB-C for our baseline). It is an isotropic ViT, so the shared head applies unchanged, and its keypoint-relative encoding adds $0.018$ GFLOPs per exit, under $0.4\%$ at the depths compared here. Its baseline collapses at early exits as the plain one does ($64.72$ and $82.96$ high-quality average at quarter and half depth, $12.67$ and $53.52$ IJB-C), while KP-RPE with SD reaches $94.11$ and $96.56$ there against $93.68$ and $96.42$ for SD-ViT on the plain backbone, and $88.19$ and $95.27$ IJB-C against $87.01$ and $95.11$. At full depth it loses $0.09$ on the high-quality average and $0.36$ on IJB-C.

\textbf{Where and how many exits to supervise.}
Would fixing the auxiliary exit at one or two depths suffice? Fixing yields a sharply peaked profile (Fig.~\ref{fig:ablations}(d)): supervising only block $6$ gives $17.9\%$ and $38.1\%$ IJB-C at exits $2$ and $3$ against $74.4\%$ and $87.0\%$ for sampling, which instead spreads supervision over all depths. The fixed variants are better \emph{at} their chosen depths: $\{6\}$ reaches $96.07\%$ at exit $6$ against $95.11\%$ for sampling, and $\{3,6\}$ reaches $91.89\%$ against $87.01\%$ at exit $3$. Under the gallery-fixed protocol $\{3,6\}$ also beats sampling at its two depths ($94.33$ and $95.97$ against $93.06$ and $95.84$). The trade is therefore up to five points at the supervised depths. Away from those depths the fixed variants fall behind ($48.7\%$ at exit $2$ for $\{3,6\}$ against $74.4\%$). Committing to $\{3,6\}$ is the right choice only if those two budgets are known before training and never change.

Fixed variants are expected to fail where they are not supervised, so they do not by themselves justify sampling. The relevant comparison is \emph{deterministic} supervision of all $L{-}1$ intermediate exits, the upper bound sampling estimates. Tab.~\ref{tab:mechanism} puts the two within half a point of each other ($86.54$/$95.15$ IJB-C against $87.01$/$95.11$, $92.73$/$95.79$ against $93.06$/$95.84$ gallery-fixed). All-exit supervision costs $88.3$ A100 GPU-h against $68.6$, i.e.\ $7.4\times$ the \emph{auxiliary} cost over the $65.5$ GPU-h baseline ($22.8$ versus $3.1$). The accuracy therefore comes from shared-classifier deep supervision itself, and sampling is a single-sample, constant-cost estimator of it whose contribution is efficiency.

\textbf{What has to be shared.}
Tab.~\ref{tab:mechanism} varies the components of Eq.~\ref{eq:loss} one at a time. Blocks (i) and (ii) are strict controls, each one run from scratch at SD-ViT's setting except for the marked component. Block (iii) cannot be matched that way, since freezing prototypes at a full-depth solution and fitting per-exit heads on a frozen backbone both presuppose a model trained beforehand. First, the auxiliary exit must be supervised by the margin loss itself: replacing it with a feature-matching MSE onto the model's own final embedding, keeping the shared head, costs $31.5$ points at quarter depth ($55.49$ vs.\ $87.01$). Second, with the head shared, a separate prototype set $W'$ for the intermediate exits is nearly free ($86.90$ vs.\ $87.01$ self, $91.76$ vs.\ $93.06$ gallery-fixed), so sharing the centers adds about a point. From block (iii), per-exit heads with per-exit prototypes lose the property: ViT-FREE$_{FT}$ falls to $61.78$ gallery-fixed at quarter depth against $93.06$, and to $16.93$ against $59.31$ on TinyFace. It changes the head, the prototypes and the training regime together, so it does not isolate the head. The controls thus establish that the sampled exit needs the margin loss itself.
Freezing $W$ at a separately trained full-depth solution improves SD-ViT's quarter-depth exit ($89.93$ vs.\ $87.01$ self, $93.42$ vs.\ $93.06$ gallery-fixed). Without the sampled exit, the frozen centers alone reach $5.81$ and $76.78$ gallery-fixed at exits 3 and 6, against the baseline's $0.02$ and $34.63$. Frozen centers thus pull intermediate depths toward the gallery's space only weakly ($18.48$ self at quarter depth), and help most combined with the sampled exit.

\begin{table}[tb]
\centering
\caption{Factorized controls (ViT-S/12, WF4M, IJB-C TAR@FAR$=10^{-4}$). Blocks (i) and (ii) are strict controls: one run from scratch with the backbone, data, schedule and seed of SD-ViT, changing only the marked component. Block (iii) needs a full-depth model trained beforehand, and its GPU-h count only the second stage. \emph{shared}: the final exit's. \emph{separate}: one extra set shared by all intermediate exits. \emph{per-exit}: one per depth. \emph{frozen}: fixed at the prior full-depth solution. \emph{none}: the auxiliary exit is trained by feature-MSE onto the model's own final embedding and sees no prototypes. \#s: intermediate exits supervised per step. \emph{self}: each exit against itself. \emph{gf}: gallery-fixed (Sec.~\ref{sec:gallery}).}
\label{tab:mechanism}
\scriptsize
\scriptsize
\setlength{\tabcolsep}{4pt}
\resizebox{\linewidth}{!}{%
\begin{tabular}{l|ccc|c|cc|cc}
\hline
 & & & & & \multicolumn{2}{c|}{self} & \multicolumn{2}{c}{gf} \\
Variant & $\mathcal{H}$ & $W$ & \#s & GPU-h & ex.3 & ex.6 & ex.3 & ex.6 \\
\hline
\multicolumn{9}{l}{\emph{(i) One run from scratch, what the two exits share}} \\
Baseline ($\omega{=}1$) & \multicolumn{2}{c}{\emph{no aux.\ exit}} & 0 & 65.5 & 11.05 & 42.22 & 0.02 & 34.63 \\
aux.\ exit by feature-MSE & shared & none & 1 & 68.4 & 55.49 & 85.23 & 71.16 & 90.69 \\
separate intermediate $W'$ & shared & separate & 1 & 66.6 & 86.90 & 94.59 & 91.76 & 95.11 \\
\textbf{SD-ViT} & shared & shared & 1 & 68.6 & \textbf{87.01} & \textbf{95.11} & \textbf{93.06} & \textbf{95.84} \\
\hline
\multicolumn{9}{l}{\emph{(ii) One run from scratch, which exits are supervised}} \\
fixed exit $\{6\}$ & shared & shared & 1 & 69.2 & 38.13 & \textbf{96.07} & 40.26 & \textbf{96.31} \\
fixed exits $\{3,6\}$ & shared & shared & 2 & 72.5 & \textbf{91.89} & 95.57 & \textbf{94.33} & 95.97 \\
all exits, deterministic & shared & shared & 11 & 88.3 & 86.54 & 95.15 & 92.73 & 95.79 \\
\hline
\multicolumn{9}{l}{\emph{(iii) Consumes a separately trained full-depth model,}} \\
\multicolumn{9}{l}{\emph{\phantom{(iii) }GPU-h below exclude that model's $65.5$}} \\
ViT-FREE$_{FT}$ \cite{vitfree} & per-exit & per-exit & 1 & 40.3 & 71.36 & 89.74 & 61.78 & 89.09 \\
KD, $W$ frozen at full depth & shared & frozen & 0 & 64.2 & 18.48 & 67.97 & 5.81 & 76.78 \\
\textbf{SD-ViT}, $W$ frozen at full depth & shared & frozen & 1 & 68.8 & \textbf{89.93} & \textbf{95.93} & \textbf{93.42} & \textbf{96.23} \\
\hline
\end{tabular}}
\end{table}

\textbf{Comparison with knowledge distillation.}
Tab.~\ref{tab:distill} compares SD-ViT with distillation baselines, with and without an external teacher. Each replaces the shared-classifier coupling with a feature-matching term $\lambda\|\hat{e}^{\mathrm{s}}-\hat{e}^{\mathrm{t}}\|_2^2$ ($\lambda{=}100$) added to $\mathcal{L}_{\mathrm{cls}}(e^{\mathrm{p}},y)$ \cite{FitNet,DBLP:journals/access/BoutrosSKDKK22}, with the teacher $e^{\mathrm{t}}$ either the model's own final embedding (\emph{Feature-MSE}, the variant analysed in Sec.~\ref{sec:gallery}) or that of the separately trained, frozen Baseline (\emph{Feature-MSE (teacher)}). \emph{KD, teacher MSE} applies the same term to the final embedding $\hat{e}^{\mathrm{p}}$ instead. A prototype-level variant initialises $W$ from that full-depth model and freezes it (\emph{frozen full-depth $W$}), with or without the sampled exit.

Every baseline lifts early exits above the no-auxiliary baseline yet falls short of SD-ViT. At quarter depth SD-ViT reaches $87.0\%$ IJB-C against $55.5\%$ for the best feature-MSE variant and below $22\%$ for external-teacher distillation. The shared \emph{classifier}, not feature matching, is what makes early exits work, and an external teacher is unnecessary: prototypes frozen at a separately trained full-depth solution shift quarter-depth IJB-C by under $3$ points, while a teacher-to-intermediate MSE is worse than the model's own target ($40.3\%$ vs.\ $55.5\%$). All methods converge at full depth, and SD-ViT trains more cheaply than the teacher-MSE variants, which need a second forward pass ($68.6$ vs.\ $84.7$/$85.7$ GPU-h).

\begin{table}[t]
\centering
\caption{Auxiliary objectives (ViT-S/12, WebFace4M). Accuracy at quarter (exit 3, $4.70$ GFLOPs), half (exit 6, $9.26$ GFLOPs), and full depth (exit 12, $18.37$ GFLOPs). All reach comparable full-depth accuracy and differ almost entirely at early exits.}
\label{tab:distill}
\scriptsize
\setlength{\tabcolsep}{4pt}
\resizebox{\linewidth}{!}{%
\begin{tabular}{l|cc|ccc|ccc}
\hline
 & Ext. & Early- & \multicolumn{3}{c|}{IJB-C (TAR@$10^{-4}$)} & \multicolumn{3}{c}{TinyFace R-1} \\
Method & teacher & exit & exit 3 & exit 6 & exit 12 & exit 3 & exit 6 & exit 12 \\
\hline
Baseline (Teacher) & & & 11.05 & 42.22 & 96.89 & 40.61 & 56.01 & \textbf{74.25} \\
\hline
KD, teacher MSE & \checkmark & & 21.48 & 68.11 & \textbf{96.98} & 46.59 & 60.03 & 73.50 \\
KD, frozen full-depth $W$ & \checkmark & & 18.48 & 67.97 & 96.94 & 43.19 & 61.88 & 73.55 \\
\hline
Feature-MSE & & \checkmark & 55.49 & 85.23 & 96.82 & 53.00 & 63.36 & 73.77 \\
\textbf{SD-ViT} & & \checkmark & 87.01 & 95.11 & 96.51 & \textbf{64.62} & \textbf{71.22} & 73.85 \\
\hline
Feature-MSE (teacher) & \checkmark & \checkmark & 40.26 & 82.45 & 96.86 & 52.52 & 63.47 & 73.44 \\
\textbf{SD-ViT}, frozen full-depth $W$ & \checkmark & \checkmark &\textbf{89.93} & \textbf{95.93} & 96.68 & 62.98 & 70.98 & 73.23 \\
\hline
\end{tabular}}
\end{table}

\subsection{Gallery Consistency}
\label{sec:gallery}

An early-exit probe should match a gallery enrolled \emph{once at full depth}. The gallery-fixed protocol embeds one element of each pair at the final block and the other at exit $k$, so accuracy measures cross-depth matching. We evaluate the baseline, Feature-MSE, ViT-FREE$_{FT}$ and SD-ViT.

\begin{table}[t]
\centering
\caption{Gallery-fixed verification (ViT-S/12, WebFace4M): an exit-$k$ probe matched against a gallery enrolled once at the final block, on the five high-quality sets (average), TinyFace, and IJB-C. Unlike the self-evaluation of Tab.~\ref{tab:main_results}, this is the single-full-depth-gallery deployment setting. SD-ViT stays usable from exit 3 on, the baseline collapses, and the MSE variant lies between. ViT-FREE$_{FT}$ was not designed for this protocol (Sec.~\ref{sec:related}).}
\label{tab:gallery}
\scriptsize
\setlength{\tabcolsep}{3.6pt}
\resizebox{0.8\linewidth}{!}{%
\begin{tabular}{l|ccccc}
\hline
 & \multicolumn{5}{c}{Probe exit $k$ (vs.\ final-block gallery)} \\
Method & 1 & 3 & 6 & 9 & 12 \\
\hline
\multicolumn{6}{l}{\emph{HQ\,(5) average accuracy}} \\
Baseline & 61.07 & 66.78 & 88.56 & 96.54 & \textbf{97.43} \\
Feature-MSE      & 79.86 & 91.20 & 95.44 & 96.95 & 97.41 \\
ViT-FREE$_{FT}$ \cite{vitfree} & 86.97 & 89.55 & 95.09 & 96.95 & 97.16 \\
SD-ViT (ours)          & \textbf{90.98} & \textbf{96.15} & \textbf{97.00} & \textbf{97.25} & 97.29 \\
\hline
\multicolumn{6}{l}{\emph{TinyFace Rank-1}} \\
Baseline & 0.00 & 0.00 & 0.75 & 60.17 & \textbf{74.25} \\
Feature-MSE      & 2.39 & 24.97 & 52.07 & 68.88 & 73.77 \\
ViT-FREE$_{FT}$ \cite{vitfree} & 8.99 & 16.93 & 52.36 & 70.04 & 72.96 \\
SD-ViT (ours)          & \textbf{14.00} & \textbf{59.31} & \textbf{71.06} & \textbf{73.63} & 73.85 \\
\hline
\multicolumn{6}{l}{\emph{IJB-C TAR@FAR$=10^{-4}$}} \\
Baseline & 0.05 & 0.02 & 34.63 & 92.35 & \textbf{96.89} \\
Feature-MSE     & 16.33 & 71.16 & 90.69 & 95.50 & 96.82 \\
ViT-FREE$_{FT}$ \cite{vitfree} & 47.68 & 61.78 & 89.09 & 95.75 & 96.51 \\
SD-ViT (ours)          & \textbf{64.27} & \textbf{93.06} & \textbf{95.84} & \textbf{96.43} & 96.51 \\
\hline
\end{tabular}}
\end{table}

\textbf{Embeddings align across depth.}
Fig.~\ref{fig:main_results}(h) gives the mean cosine similarity between each intermediate embedding and the final one. The baseline's early embeddings are nearly orthogonal to it at exit $3$ ($0.14$) and still far from it at exit $6$ ($0.40$). SD-ViT raises this to $0.76$ and $0.91$, more than the explicit MSE objective ($0.41$ and $0.64$).

\textbf{Recognition against a full-depth gallery.}
Fig.~\ref{fig:main_results}(e-g) and Tab.~\ref{tab:gallery} report recognition against a fixed full-depth gallery. On the high-quality sets the baseline manages $66.8\%$ at quarter depth against SD-ViT's $96.2\%$, and $91.0\%$ even at the first exit. The gap widens on harder benchmarks: the baseline is unusable below half depth ($0.0\%$ TinyFace and $0.0\%$ IJB-C at quarter depth) where SD-ViT reaches $59.3\%$ and $93.1\%$.

\textbf{One threshold across all depths.}
TAR at a target FAR is a ROC-level statistic: it permits a different threshold at every exit, so Tab.~\ref{tab:gallery} alone does not show that a deployed system can keep its threshold when it starts exiting early. We split each benchmark's pairs once into a development and a test half, choose a single $\tau$ on the development half of the \emph{full-depth} scores at the target FAR, and apply it unchanged at every exit on the test half. Tab.~\ref{tab:calib} reports the realized FAR and TAR against a $\tau$ re-calibrated per exit. On IJB-C at FAR$=10^{-4}$, $\tau{=}0.382$ costs SD-ViT $0.19$ points of TAR at exit $9$, $0.64$ at exit $6$ and $3.97$ at quarter depth, so half depth needs no recalibration. The baseline cannot be operated this way. It gets nothing at exits $1$ and $3$, $19.59\%$ against $34.19\%$ with a per-exit $\tau$ at exit $6$, and becomes operable only by exit~$9$ ($90.61$ against $92.16$). Its usable range under a fixed threshold thus begins three quarters of the way down the network. Realized FAR shrinks as the probe gets shallower and stays below the target at every early exit. At exit $1$, however, even SD-ViT keeps only $15.66\%$ against $63.81\%$ with a per-exit $\tau$. The single-threshold property therefore holds from quarter depth onward, not at every exit. We ran this calibration on ViT-S/12 only.

\begin{table}[t]
\centering
\caption{One threshold, calibrated once at full depth, held fixed across exits (ViT-S/12, WF4M, gallery-fixed protocol). $\tau$ is chosen on a development half of the pairs so that the full-depth FAR equals the target. All rows then apply that same $\tau$ on the held-out half, so the full-depth FAR deviates from the target only through the split. \emph{per-exit $\tau$} re-calibrates $\tau$ on each exit's own development scores and is an upper bound that a deployed system cannot use.}
\resizebox{0.8\linewidth}{!}{%
\label{tab:calib}
\scriptsize
\setlength{\tabcolsep}{3.4pt}
\begin{tabular}{l|c|cc|c}
\hline
 & Probe & \multicolumn{2}{c|}{fixed $\tau$ (full depth)} & per-exit $\tau$ \\
Method & exit $k$ & FAR$\times10^{4}$ & TAR & TAR \\
\hline
\multicolumn{5}{l}{\emph{IJB-C, target FAR$=10^{-4}$, $\tau$ fixed at full depth}} \\
Baseline           & 3  & 0.03 & 0.00  & 0.01  \\
                   & 6  & 0.12 & 19.59 & 34.19 \\
                   & 9  & 0.41 & 90.61 & 92.16 \\
                   & 12 & 0.96 & 96.91 & 96.91 \\
\hline
Feature-MSE  & 3  & 0.00 & 29.24 & 70.80 \\
                   & 6  & 0.01 & 79.56 & 90.48 \\
                   & 9  & 0.31 & 94.36 & 95.47 \\
                   & 12 & 0.98 & 96.78 & 96.78 \\
\hline
SD-ViT (ours)      & 3  & 0.13 & \textbf{89.08} & 93.05 \\
                   & 6  & 0.43 & \textbf{95.15} & 95.78 \\
                   & 9  & 0.75 & \textbf{96.15} & 96.34 \\
                   & 12 & 1.02 & 96.48 & 96.48 \\
\hline
\end{tabular}}
\end{table}

\section{Conclusion}
We presented SD-ViT, a stochastic shared-classifier deep-supervision framework that turns a standard ViT into a face recognizer with a usable exit at every block depth. A uniformly sampled intermediate block is decoded by the same head and scored against the same prototypes as the final layer, with no teacher and no per-exit adaptation layers. An early-exit probe therefore matches a gallery enrolled once at full depth by plain cosine similarity, with a single verification threshold held fixed from quarter depth onward. Across ViT-S and ViT-B on WebFace4M and WebFace12M, SD-ViT stays within $0.4$ IJB-C points of the identically trained baseline at full depth while recovering over $96\%$ of its own full-depth accuracy at half the computation. Our controls show that the sampled exit needs the margin loss itself, not feature matching. Future work includes input-adaptive per-image exit policies.

\section*{Acknowledgment}
This research work has been funded by the German Federal Ministry of Education and Research and the Hessen State Ministry for Higher Education, Research and the Arts within their joint support of the National Research Center for Applied Cybersecurity ATHENE.

\clearpage
{
    \small
    \bibliographystyle{ieeenat_fullname}
    \bibliography{main}
}

\clearpage
\appendix

% S-prefixed numbering so this document does not clash with the main paper
\renewcommand{\thesection}{S\arabic{section}}
\renewcommand{\thesubsection}{S\arabic{section}.\arabic{subsection}}
\renewcommand{\thetable}{S\arabic{table}}
\renewcommand{\thefigure}{S\arabic{figure}}
\renewcommand{\theequation}{S\arabic{equation}}
\setcounter{section}{0}\setcounter{table}{0}\setcounter{figure}{0}\setcounter{equation}{0}

\begin{Large}
\noindent \textbf{Supplementary Material}\\
\end{Large}

The main paper mostly presents the quarter-, half-, and full-depth operating points and the \emph{average} of the five high-quality verification sets, whereas here we give per-exit results on the individual benchmarks, together with the cost and parameter accounting, the training budget of the runs, and the per-dataset gallery-consistency analysis.

\textbf{Notation.} Throughout, ``Exit $k$'' denotes terminating inference after the first $k$ transformer blocks and decoding the resulting tokens with the shared embedding head (Sec.~3 of the main paper). Unless noted otherwise, all per-exit results use the self-evaluation protocol (probe and gallery embedded at the same exit), the gallery-fixed protocol of Sec.~\ref{sec:supp_gallery} instead matches an exit-$k$ probe against a final-block gallery. ``HQ\,(5)'' is the average of LFW, AgeDB-30, CFP-FP, CPLFW, and CALFW.

\section{Computational Cost and Training Budget}
\label{sec:supp_cost}

\textbf{Computational cost.} Table~\ref{tab:cost} lists the exact per-exit cost and the size of the network retained when the model is pruned at each exit.

\textbf{Training budget.} Table~\ref{tab:gpu} lists the wall-clock training cost of every run in A100 GPU-hours. The stochastic auxiliary loss term processes a single additional intermediate exit per step and adds only a few percent of overhead, it adds \emph{no} inference-time cost, since the deployed network is architecturally identical to the baseline.

\section{Full Per-Exit Results on WebFace4M}
\label{sec:supp_main}

Tables~\ref{tab:supp_s12} and \ref{tab:supp_b24} give the complete per-exit breakdown for the two backbones of Fig.~2 (top row) and Tab.~1, on all eight evaluation protocols: the five high-quality verification sets individually (and their average), TinyFace Rank-1 and Rank-5, and IJB-C at all three operating points ($10^{-5}$, $10^{-4}$, $10^{-3}$). The pattern summarized in the main paper holds on every benchmark: the baseline degrades severely as the exit is moved earlier, while SD-ViT degrades far less (ViT-S/12 IJB-C $40.32$ at exit~1 and $87.01$ at exit~3, against $96.51$ at full depth). The gap is largest at the early exits (up to $76.0$ IJB-C points, at exit~3 of ViT-S/12) and closes near full depth, where the baseline is slightly ahead (from exit~11 of ViT-S/12 and exit~21 of ViT-B/24).

\section{Scaling to WebFace12M}
\label{sec:supp_wf12m}

Table~\ref{tab:supp_wf12m} gives the full per-exit breakdown for ViT-B/24 trained on the larger WebFace12M. Larger-scale training improves both models at full depth, but does not by itself make the baseline's intermediate layers usable: at quarter depth the WF12M baseline still collapses (e.g.\ $12.1\%$ TAR at $10^{-4}$ on IJB-C), while SD-ViT reaches $95.0\%$ there.

\section{Loss-Function Ablation: ArcFace}
\label{sec:supp_arcface}

We repeat the ViT-S/12 experiment with ArcFace in place of AdaFace. Table~\ref{tab:supp_arc} gives the full per-exit breakdown for both the ArcFace baseline and ArcFace SD-ViT. The pattern matches the AdaFace case: SD-ViT lifts the early exits far above the baseline ($80.60$ vs.\ $61.04$ HQ\,(5) at exit~1, $93.71$ vs.\ $68.04$ at exit~3) and stays within $0.3$ points of it at full depth ($97.20$ vs.\ $97.41$ HQ\,(5), $96.65$ vs.\ $96.93$ IJB-C), with the first two exits still short of full depth as on AdaFace.

\section{Primary-Loss Weight}
\label{sec:supp_omega}

Tables~\ref{tab:supp_w00}-\ref{tab:supp_w09} give the full per-exit breakdown for the remaining values of the primary-loss weight $\omega$ studied in Fig.~3(a) and Tab.~2, the chosen $\omega{=}0.7$ run is Table~\ref{tab:supp_s12} (SD-ViT block) and $\omega{=}1.0$ is the baseline. Increasing $\omega$ trades early-exit accuracy for full-depth accuracy (full-depth IJB-C $96.20$, $96.51$ and $96.73$ at $\omega{=}0.5$, $0.7$ and $0.9$, against $96.89$ for the baseline). $\omega{=}0.0$, which supervises only the sampled early exit, keeps early exits strong but lowers full-depth IJB-C to $95.25$.

\section{Network Depth: ViT-S/15}
\label{sec:supp_depth}

Table~\ref{tab:supp_d15} gives the full per-exit breakdown for the deeper ViT-S/15 backbone (the mid- and full-depth summary is in Sec.~4.3 of the main paper).

\section{Where to Supervise}
\label{sec:supp_where}

Section~4.3 of the main paper argues that stochastically sampling the auxiliary exit is preferable to supervising one or two fixed depths. Tables~\ref{tab:supp_e36} and \ref{tab:supp_e6} give the full per-exit breakdown of the two fixed-exit variants (the stochastic counterpart is Table~\ref{tab:supp_s12}, SD-ViT block). Supervising at fixed exits produces a sharply peaked profile that is strong at and after the chosen depth but collapses earlier, most visibly for the exit-$\{6\}$ variant, whose IJB-C TAR at exit~2 is only $17.9\%$ versus $74.4\%$ for stochastic sampling. The fixed variants are stronger \emph{at} their chosen depths: the $\{3,6\}$ variant reaches $91.89\%$ IJB-C at exit~3 against SD-ViT's $87.01\%$, and under the gallery-fixed protocol, reported in Table~3 of the main paper, it supervises both of its depths better than sampling does ($94.33$ and $95.97$ against $93.06$ and $95.84$).

\section{Auxiliary Objectives}
\label{sec:supp_mse}

Sec.~4.3 and Tab.~4 of the main paper compare SD-ViT against auxiliary objectives along two axes: whether a method uses an external pretrained teacher, and whether it targets early-exit usability. This section gives the per-exit breakdown behind those two columns: external-teacher variants in Table~\ref{tab:supp_kd}, feature-matching variants in Table~\ref{tab:supp_selfmse}, and the two that bring the teacher to the sampled exit in Table~\ref{tab:supp_sdteacher}. The Baseline and SD-ViT rows are in Table~\ref{tab:supp_s12}. All runs use ViT-S/12 on WebFace4M with AdaFace. Every feature-matching baseline replaces the shared-classifier coupling by

\begin{equation}
\label{eq:supp_mse}
\mathcal{L} = \mathcal{L}_{\mathrm{cls}}(e^{\mathrm{p}}, y) + \lambda \big\| \hat{e}^{\mathrm{s}} - \hat{e}^{\mathrm{t}} \big\|_2^2, \qquad \lambda = 100,
\end{equation}
with the margin loss applied to $e^{\mathrm{p}}$ only and $\hat{e}^{\mathrm{t}}$ either the model's own final embedding (\emph{Feature-MSE}) or that of the Baseline, trained separately and frozen (\emph{Feature-MSE (teacher)}). In both, the sampled exit receives no label loss. \emph{KD, teacher MSE} applies the same term to the final embedding, with $\hat{e}^{\mathrm{p}}$ in place of $\hat{e}^{\mathrm{s}}$, i.e.\ standard embedding distillation into a single-exit student. The prototype-level variant instead freezes $W$ at the separately trained full-depth solution (\emph{frozen full-depth $W$}). Table~\ref{tab:supp_selfmse} additionally separates two placements of the MSE that the main paper does not distinguish: \emph{pre-proj.} matches the token sequences before $\phi$, and \emph{post-proj.} matches the embeddings after it. The post-projection variant is the \emph{Feature-MSE} of the main paper and the one analyzed in Sec.~\ref{sec:supp_gallery}. Matching before the projection is worse in $96$ of the $99$ per-exit entries at exits 1 to 9 (the exceptions are CFP-FP and IJB-C at $10^{-3}$ at exit~7 and TinyFace R-5 at exit~9), by up to $38.5$ IJB-C points (exit~4), and the two placements are within about a point of each other from exit~10 on.

\section{Gallery Consistency: Full Results}
\label{sec:supp_gallery}

Tables~\ref{tab:gf_full} and \ref{tab:gf_hard} give the full per-exit gallery-fixed results behind Sec.~4.4 and Tab.~5: an exit-$k$ probe matched against a gallery embedded at the final block. A value of $0.00$ is a genuine collapse, not a missing entry.

\section{Embedding Alignment Across Depth}
\label{sec:supp_sim}

Tables~\ref{tab:sim_ViT-S}-\ref{tab:sim_ViT-SSD07} give the per-dataset cosine similarity between each intermediate embedding and the final-block embedding, for each of the three models and each of the seven test sets. The main paper plots their average (Fig.~2(h)). SD-ViT has the highest similarity on every dataset and at every exit. The baseline's early embeddings are nearly orthogonal to its final embedding (at most $0.26$ up to exit~3), and the explicit-MSE variant lies between the two, except on TinyFace at exits~1, 9 and 10, where it is below the baseline. At exit~11 all three are within $0.04$.

\section{Comparison with ViT-FREE and Matched-Compute Models}
\label{sec:supp_vitfree}

Sec.~2 and Sec.~4.1 of the main paper explain why training-free ViT-FREE coincides with our Baseline read-out, and Tab.~1 and Tab.~5 give its headline numbers. ViT-FREE$_{FT}$ follows the original method: one projection head per exit, fine-tuned with its own classifier while the backbone stays frozen. For a like-for-like comparison, the backbone is our ViT-S/12 Baseline and the heads are trained with its data and loss (WebFace4M, AdaFace), at exits $\{1,3,6,9,12\}$. Table~\ref{tab:supp_vitfree} reports both protocols at those exits. Table~\ref{tab:supp_trunc} lists the matched-compute references discussed in Sec.~4.2: ViT-S backbones of $3$ and $6$ blocks trained from scratch with the identical setting.

\section{Deterministic Supervision of All Exits}
\label{sec:supp_allexit}

Sec.~4.3 compares stochastic sampling against deterministic supervision of all $L{-}1$ intermediate exits, loss $=0.7\,\mathcal{L}(e^{(L)})+0.3\cdot \frac{1}{L-1}\sum_{k<L}\mathcal{L}(e^{(k)})$, with the same shared head, prototypes and $\omega$. Table~\ref{tab:supp_allexit} gives that run per exit under both protocols.

\section{Factorized Mechanism Controls}
\label{sec:supp_mechanism}

Table~3 of the main paper is summarized at two exits. Table~\ref{tab:supp_splitproto} gives the separate-intermediate-$W'$ control per exit under both protocols. Per-exit heads are covered only by ViT-FREE$_{FT}$ (Sec.~\ref{sec:supp_vitfree}), which changes the head and the prototypes together.

\section{One Threshold Across Depths}
\label{sec:supp_calib}

\begin{figure*}[t]
\centering
\includegraphics[width=\linewidth]{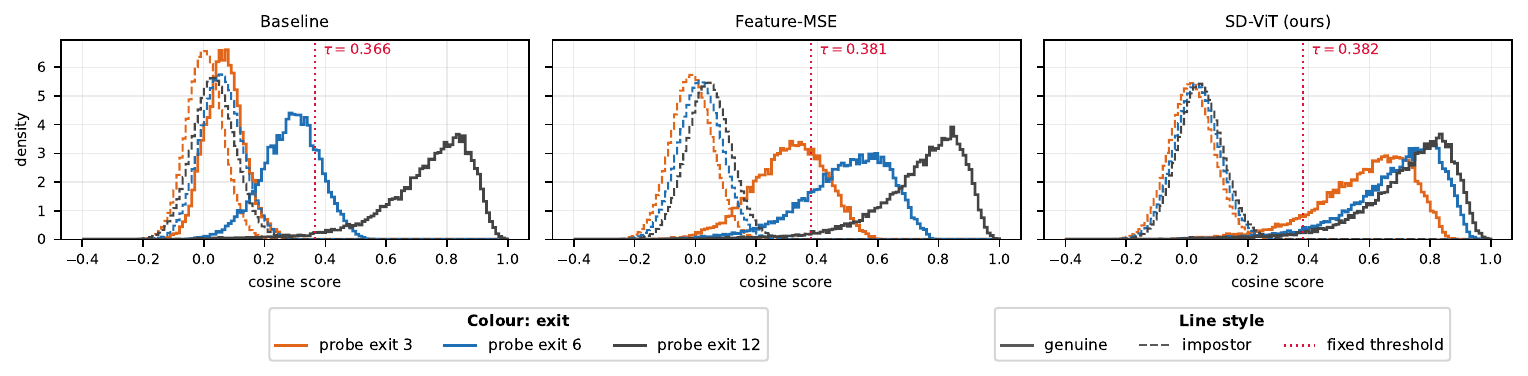}
\caption{Genuine and impostor score distributions on IJB-C under the gallery-fixed protocol, for probes taken at exits 3, 6 and 12 against a gallery enrolled once at full depth (ViT-S/12, WF4M). The dotted line is the single threshold $\tau$ calibrated on the full-depth development half at FAR$=10^{-4}$ and then held fixed. For the Baseline the genuine scores at exit 3 lie almost on top of the impostor distribution and essentially none of them clear $\tau$. At exit 6 they have moved up but only about a fifth clear it ($19.59\%$ TAR, Tab.~6). For SD-ViT the genuine distributions of all three exits are close to superimposed and sit well above $\tau$, while the impostor distributions stay centred near zero at every depth. The residual left-shift of the exit-3 genuine mode is what costs the $3.97$ points of TAR reported in Sec.~4.4 of the main paper.}
\label{fig:supp_scores}
\end{figure*}

Sec.~4.4 defines the protocol: one $\tau$ calibrated on a development half of the \emph{full-depth} scores and applied unchanged at every exit on the held-out half. Table~\ref{tab:supp_calib} extends Tab.~6 to exit~1 and lists the realized FAR and the mean genuine and impostor scores, and Fig.~\ref{fig:supp_scores} shows the score distributions behind it. The calibration was run on ViT-S/12 only.

\section{Measured Latency and Memory}
\label{sec:supp_latency}

Table~\ref{tab:cost} reports estimated GFLOPs and parameter counts. Because a FLOP count need not translate into wall-clock savings, Table~\ref{tab:supp_lat} adds measured batch-1 latency and peak memory per exit on named hardware, for all three configurations. SD-ViT and the Baseline share an architecture, so these numbers apply to both.

\section{A Second ViT Backbone: KP-RPE}
\label{sec:supp_archs}

The objective requires a backbone whose token dimensionality is constant across depth, so that one head can decode every exit (Sec.~3.2 of the main paper). This section reports a second backbone that satisfies it. Sec.~4.3 reports the KP-RPE result, Table~\ref{tab:supp_kprpe} gives it per exit. Its GFLOPs are measured for this backbone, not copied from the plain ViT: the keypoint-relative encoding projects the five keypoints into the per-block, per-head attention biases once per image, adding $0.018$ GFLOPs at every exit, $+1.1\%$ at exit~1 falling to $+0.1\%$ at full depth.

\clearpage

\begin{table}[t]
\centering
\caption{Per-exit computational cost. For each exit $k$ we report inference cost (GFLOPs) and the number of parameters of the network retained when the model is truncated at that exit (\#Par, in millions). ViT-S columns cover both ViT-S/12 and ViT-S/15 (identical small configuration), ViT-B columns are the base configuration. Full-depth is $k{=}12$ (ViT-S/12), $k{=}15$ (ViT-S/15), and $k{=}24$ (ViT-B/24). - marks a depth the backbone does not have.}
\label{tab:cost}
\footnotesize
\setlength{\tabcolsep}{4pt}
\begin{tabular}{c|cc|cc}
\hline
 & \multicolumn{2}{c|}{ViT-S (Small)} & \multicolumn{2}{c}{ViT-B (Base)} \\
Exit $k$ & GFLOP & \#Par (M) & GFLOP & \#Par (M) \\
\hline
1 & 1.66 & 55.5 & 1.25 & 54.5 \\
2 & 3.18 & 59.2 & 2.36 & 57.1 \\
3 & 4.70 & 62.9 & 3.47 & 59.7 \\
4 & 6.22 & 66.5 & 4.58 & 62.4 \\
5 & 7.74 & 70.2 & 5.68 & 65.0 \\
6 & 9.26 & 73.9 & 6.79 & 67.6 \\
7 & 10.78 & 77.6 & 7.90 & 70.2 \\
8 & 12.30 & 81.2 & 9.01 & 72.8 \\
9 & 13.82 & 84.9 & 10.12 & 75.5 \\
10 & 15.34 & 88.6 & 11.23 & 78.1 \\
11 & 16.86 & 92.3 & 12.33 & 80.7 \\
12 & 18.37 & 96.0 & 13.44 & 83.4 \\
13 & 19.89 & 99.6 & 14.55 & 86.0 \\
14 & 21.41 & 103.3 & 15.66 & 88.6 \\
15 & 22.93 & 107.0 & 16.77 & 91.2 \\
16 & - & - & 17.88 & 93.9 \\
17 & - & - & 18.98 & 96.5 \\
18 & - & - & 20.09 & 99.1 \\
19 & - & - & 21.20 & 101.7 \\
20 & - & - & 22.31 & 104.4 \\
21 & - & - & 23.42 & 107.0 \\
22 & - & - & 24.53 & 109.6 \\
23 & - & - & 25.63 & 112.2 \\
24 & - & - & 26.74 & 114.9 \\
\hline
\end{tabular}
\end{table}

\begin{table}[t]
\centering
\caption{Training cost of every run, in NVIDIA A100 GPU-hours. The stochastic auxiliary exit adds only a few percent over the corresponding baseline (e.g.\ $65.5\!\to\!68.6$ for ViT-S/12 and $411.1\!\to\!430.2$ for ViT-B/24 on WF12M). Sampling one exit per step costs $3.1$ GPU-h over the baseline, supervising all eleven costs $22.8$.}
\label{tab:gpu}
\footnotesize
\setlength{\tabcolsep}{4pt}
\resizebox{\linewidth}{!}{%
\begin{tabular}{l|llcc|c}
\hline
Method & Backbone & Data & Loss & $\omega$ & GPU-h \\
\hline
Baseline & ViT-S/12 & WF4M & AdaFace & - & 65.5 \\
SD-ViT & ViT-S/12 & WF4M & AdaFace & 0.7 & 68.6 \\
Baseline & ViT-S/12 & WF4M & ArcFace & - & 65.3 \\
SD-ViT & ViT-S/12 & WF4M & ArcFace & 0.7 & 68.8 \\
SD-ViT & ViT-S/12 & WF4M & AdaFace & 0.0 & 70.3 \\
SD-ViT & ViT-S/12 & WF4M & AdaFace & 0.5 & 71.2 \\
SD-ViT & ViT-S/12 & WF4M & AdaFace & 0.9 & 68.0 \\
SD-ViT, exits\,$\{3,6\}$ & ViT-S/12 & WF4M & AdaFace & 0.7 & 72.5 \\
SD-ViT, exit\,$\{6\}$ & ViT-S/12 & WF4M & AdaFace & 0.7 & 69.2 \\
Feature-MSE & ViT-S/12 & WF4M & AdaFace & - & 68.4 \\
KD, teacher MSE & ViT-S/12 & WF4M & AdaFace & - & 84.7 \\
KD, frozen full-depth $W$ & ViT-S/12 & WF4M & AdaFace & - & 64.2 \\
SD-ViT, frozen full-depth $W$ & ViT-S/12 & WF4M & AdaFace & 0.7 & 68.8 \\
Feature-MSE (teacher) & ViT-S/12 & WF4M & AdaFace & - & 85.7 \\
\hline
ViT-S/3, from scratch & ViT-S/3 & WF4M & AdaFace & - & 39.0 \\
ViT-S/6, from scratch & ViT-S/6 & WF4M & AdaFace & - & 41.2 \\
ViT-FREE$_{FT}$, per-exit heads and $W$ & ViT-S/12 & WF4M & AdaFace & - & 40.3 \\
Separate intermediate $W'$ & ViT-S/12 & WF4M & AdaFace & 0.7 & 66.6 \\
All exits, deterministic & ViT-S/12 & WF4M & AdaFace & 0.7 & 88.3 \\
KP-RPE, Baseline & ViT-S/12 & WF4M & AdaFace & - & 74.2 \\
KP-RPE $+$ SD & ViT-S/12 & WF4M & AdaFace & 0.7 & 76.5 \\
Baseline & ViT-S/15 & WF4M & AdaFace & - & 81.1 \\
SD-ViT & ViT-S/15 & WF4M & AdaFace & 0.7 & 83.5 \\
Baseline & ViT-B/24 & WF4M & AdaFace & - & 138.2 \\
SD-ViT & ViT-B/24 & WF4M & AdaFace & 0.7 & 143.9 \\
Baseline & ViT-B/24 & WF12M & AdaFace & - & 411.1 \\
SD-ViT & ViT-B/24 & WF12M & AdaFace & 0.7 & 430.2 \\
\hline
\end{tabular}}
\end{table}

\begin{table*}[t]
\centering
\caption{Full per-exit results for \textbf{ViT-S/12} on WebFace4M (AdaFace). Baseline is standard final-exit-only training, SD-ViT uses $\omega{=}0.7$.}
\label{tab:supp_s12}
\scriptsize
\setlength{\tabcolsep}{2.3pt}
\begin{tabular}{l c r|cccccc|cc|ccc}
\hline
Method & Exit & GFLOP & \multicolumn{6}{c|}{High-quality verification (\%)} & \multicolumn{2}{c|}{TinyFace} & \multicolumn{3}{c}{IJB-C (TAR@FAR)} \\
& $k$ & & LFW & AgeDB & CFP-FP & CPLFW & CALFW & Avg & R-1 & R-5 & $10^{-5}$ & $10^{-4}$ & $10^{-3}$ \\
\hline
\multirow{12}{*}{Baseline} & 1 & 1.66 & 70.97 & 52.40 & 61.91 & 55.20 & 56.72 & 59.44 & 29.18 & 36.72 & 2.38 & 5.56 & 12.82 \\
 & 2 & 3.18 & 74.95 & 54.18 & 63.26 & 56.52 & 58.83 & 61.55 & 34.44 & 43.32 & 3.60 & 8.14 & 17.46 \\
 & 3 & 4.70 & 79.12 & 56.27 & 63.46 & 57.43 & 62.13 & 63.68 & 40.61 & 49.28 & 5.13 & 11.05 & 21.75 \\
 & 4 & 6.22 & 83.03 & 60.83 & 65.31 & 59.35 & 66.45 & 66.99 & 45.20 & 54.45 & 8.94 & 16.87 & 29.51 \\
 & 5 & 7.74 & 87.83 & 65.83 & 67.23 & 62.32 & 71.08 & 70.86 & 50.97 & 59.42 & 16.50 & 26.99 & 40.83 \\
 & 6 & 9.26 & 93.13 & 71.82 & 74.99 & 70.67 & 78.87 & 77.90 & 56.01 & 63.49 & 29.44 & 42.22 & 56.18 \\
\hline
\end{tabular}
\end{table*}

\begin{table*}[t]
\centering
\caption{Full per-exit results for \textbf{ViT-B/24} on WebFace4M (AdaFace).}
\label{tab:supp_b24}
\scriptsize
\setlength{\tabcolsep}{2.3pt}
\begin{tabular}{l c r|cccccc|cc|ccc}
\hline
Method & Exit & GFLOP & \multicolumn{6}{c|}{High-quality verification (\%)} & \multicolumn{2}{c|}{TinyFace} & \multicolumn{3}{c}{IJB-C (TAR@FAR)} \\
& $k$ & & LFW & AgeDB & CFP-FP & CPLFW & CALFW & Avg & R-1 & R-5 & $10^{-5}$ & $10^{-4}$ & $10^{-3}$ \\
\hline
\multirow{24}{*}{Baseline} & 1 & 1.25 & 69.20 & 53.10 & 61.87 & 55.18 & 55.78 & 59.03 & 26.80 & 35.01 & 1.99 & 4.47 & 10.77 \\
 & 2 & 2.36 & 72.15 & 52.95 & 61.97 & 54.92 & 57.93 & 59.98 & 33.45 & 42.70 & 2.54 & 5.93 & 13.79 \\
 & 3 & 3.47 & 75.90 & 55.30 & 61.61 & 55.33 & 60.65 & 61.76 & 39.46 & 49.22 & 3.74 & 8.46 & 18.46 \\
 & 4 & 4.58 & 79.32 & 57.73 & 63.10 & 56.23 & 62.68 & 63.81 & 43.40 & 53.14 & 5.04 & 11.18 & 22.65 \\
 & 5 & 5.68 & 82.67 & 60.92 & 65.91 & 58.28 & 65.55 & 66.67 & 46.41 & 55.87 & 8.05 & 16.29 & 29.52 \\
 & 6 & 6.79 & 86.02 & 62.80 & 68.70 & 60.47 & 68.78 & 69.35 & 49.73 & 58.83 & 11.67 & 21.63 & 36.77 \\
\hline
\end{tabular}
\end{table*}

\begin{table*}[t]
\centering
\caption{Full per-exit results for \textbf{ViT-B/24} on WebFace12M (AdaFace).}
\label{tab:supp_wf12m}
\scriptsize
\setlength{\tabcolsep}{2.3pt}
\begin{tabular}{l c r|cccccc|cc|ccc}
\hline
Method & Exit & GFLOP & \multicolumn{6}{c|}{High-quality verification (\%)} & \multicolumn{2}{c|}{TinyFace} & \multicolumn{3}{c}{IJB-C (TAR@FAR)} \\
& $k$ & & LFW & AgeDB & CFP-FP & CPLFW & CALFW & Avg & R-1 & R-5 & $10^{-5}$ & $10^{-4}$ & $10^{-3}$ \\
\hline
\multirow{24}{*}{Baseline} & 1 & 1.25 & 70.35 & 52.63 & 61.86 & 55.17 & 56.77 & 59.36 & 23.77 & 31.22 & 1.72 & 4.07 & 10.16 \\
 & 2 & 2.36 & 70.62 & 53.65 & 61.87 & 54.28 & 56.37 & 59.36 & 27.52 & 35.76 & 1.95 & 4.79 & 11.86 \\
 & 3 & 3.47 & 73.17 & 52.77 & 61.81 & 55.02 & 57.88 & 60.13 & 30.85 & 39.86 & 2.22 & 5.54 & 13.85 \\
 & 4 & 4.58 & 76.05 & 54.70 & 62.41 & 54.80 & 59.62 & 61.52 & 35.68 & 44.61 & 2.96 & 6.90 & 16.21 \\
 & 5 & 5.68 & 78.65 & 55.67 & 63.97 & 55.85 & 61.80 & 63.19 & 39.16 & 49.06 & 3.68 & 8.70 & 19.05 \\
 & 6 & 6.79 & 80.10 & 57.80 & 64.04 & 56.37 & 63.98 & 64.46 & 44.21 & 53.22 & 5.39 & 12.10 & 23.94 \\
\hline
\end{tabular}
\end{table*}

\begin{table*}[t]
\centering
\caption{Full per-exit results for \textbf{ViT-S/12 with ArcFace} on WebFace4M.}
\label{tab:supp_arc}
\scriptsize
\setlength{\tabcolsep}{2.3pt}
\begin{tabular}{l c r|cccccc|cc|ccc}
\hline
Method & Exit & GFLOP & \multicolumn{6}{c|}{High-quality verification (\%)} & \multicolumn{2}{c|}{TinyFace} & \multicolumn{3}{c}{IJB-C (TAR@FAR)} \\
& $k$ & & LFW & AgeDB & CFP-FP & CPLFW & CALFW & Avg & R-1 & R-5 & $10^{-5}$ & $10^{-4}$ & $10^{-3}$ \\
\hline
\multirow{12}{*}{Baseline} & 1 & 1.66 & 73.65 & 53.50 & 63.47 & 56.07 & 58.53 & 61.04 & 29.45 & 37.98 & 3.07 & 7.07 & 15.44 \\
 & 2 & 3.18 & 78.63 & 56.90 & 63.84 & 57.38 & 61.48 & 63.65 & 39.67 & 48.58 & 5.20 & 11.37 & 22.59 \\
 & 3 & 4.70 & 84.32 & 61.68 & 66.29 & 60.08 & 67.83 & 68.04 & 45.79 & 54.86 & 10.40 & 18.98 & 32.61 \\
 & 4 & 6.22 & 88.83 & 67.98 & 71.06 & 63.85 & 73.38 & 73.02 & 51.31 & 59.12 & 19.04 & 30.44 & 44.79 \\
 & 5 & 7.74 & 93.05 & 72.88 & 75.37 & 69.95 & 78.82 & 78.01 & 56.79 & 63.68 & 31.10 & 44.27 & 58.29 \\
 & 6 & 9.26 & 95.93 & 77.37 & 82.04 & 78.15 & 83.58 & 83.41 & 59.92 & 66.23 & 44.39 & 57.82 & 70.80 \\
\hline
\end{tabular}
\end{table*}

\begin{table*}[t]
\centering
\caption{Full per-exit results for SD-ViT with $\omega{=}0.0$ (only the sampled early exit is supervised, ViT-S/12, WF4M, AdaFace).}
\label{tab:supp_w00}
\scriptsize
\setlength{\tabcolsep}{2.3pt}
\begin{tabular}{l c r|cccccc|cc|ccc}
\hline
Method & Exit & GFLOP & \multicolumn{6}{c|}{High-quality verification (\%)} & \multicolumn{2}{c|}{TinyFace} & \multicolumn{3}{c}{IJB-C (TAR@FAR)} \\
& $k$ & & LFW & AgeDB & CFP-FP & CPLFW & CALFW & Avg & R-1 & R-5 & $10^{-5}$ & $10^{-4}$ & $10^{-3}$ \\
\hline
\multirow{12}{*}{$\omega{=}0.0$} & 1 & 1.66 & 98.00 & 82.52 & 81.76 & 78.97 & 89.35 & 86.12 & 55.12 & 62.07 & 49.08 & 61.66 & 74.53 \\
 & 2 & 3.18 & 99.28 & 91.10 & 92.13 & 87.80 & 93.12 & 92.69 & 62.85 & 67.54 & 74.90 & 84.81 & 91.72 \\
 & 3 & 4.70 & 99.50 & 94.07 & 96.09 & 90.90 & 94.67 & 95.05 & 67.57 & 72.18 & 86.72 & 92.21 & 95.69 \\
 & 4 & 6.22 & 99.58 & 95.37 & 97.57 & 92.47 & 95.08 & 96.01 & 69.69 & 73.79 & 90.10 & 94.37 & 96.66 \\
 & 5 & 7.74 & 99.58 & 95.63 & 97.93 & 92.77 & 95.32 & 96.25 & 70.71 & 74.65 & 91.04 & 95.00 & 97.06 \\
 & 6 & 9.26 & 99.65 & 96.37 & 98.04 & 93.02 & 95.48 & 96.51 & 70.90 & 75.24 & 91.32 & 95.21 & 97.16 \\
 & 7 & 10.78 & 99.65 & 96.45 & 98.10 & 93.08 & 95.57 & 96.57 & 71.19 & 75.19 & 91.52 & 95.28 & 97.23 \\
 & 8 & 12.30 & 99.65 & 96.53 & 98.31 & 93.05 & 95.55 & 96.62 & 71.33 & 75.30 & 91.74 & 95.32 & 97.28 \\
 & 9 & 13.82 & 99.63 & 96.33 & 98.31 & 92.97 & 95.63 & 96.57 & 71.43 & 75.19 & 91.63 & 95.25 & 97.29 \\
 & 10 & 15.34 & 99.67 & 96.48 & 98.31 & 93.03 & 95.57 & 96.61 & 71.35 & 75.32 & 91.71 & 95.26 & 97.28 \\
 & 11 & 16.86 & 99.67 & 96.48 & 98.30 & 92.97 & 95.57 & 96.60 & 71.38 & 75.30 & 91.74 & 95.26 & 97.27 \\
 & 12 & 18.37 & 99.63 & 96.50 & 97.99 & 93.37 & 95.58 & 96.61 & 71.30 & 75.11 & 91.59 & 95.25 & 97.33 \\
\hline
\end{tabular}
\end{table*}

\begin{table*}[t]
\centering
\caption{Full per-exit results for SD-ViT with $\omega{=}0.5$ (ViT-S/12, WF4M, AdaFace).}
\label{tab:supp_w05}
\scriptsize
\setlength{\tabcolsep}{2.3pt}
\begin{tabular}{l c r|cccccc|cc|ccc}
\hline
Method & Exit & GFLOP & \multicolumn{6}{c|}{High-quality verification (\%)} & \multicolumn{2}{c|}{TinyFace} & \multicolumn{3}{c}{IJB-C (TAR@FAR)} \\
& $k$ & & LFW & AgeDB & CFP-FP & CPLFW & CALFW & Avg & R-1 & R-5 & $10^{-5}$ & $10^{-4}$ & $10^{-3}$ \\
\hline
\multirow{12}{*}{$\omega{=}0.5$} & 1 & 1.66 & 97.15 & 78.20 & 78.99 & 76.15 & 87.75 & 83.65 & 51.88 & 58.34 & 33.88 & 47.65 & 61.29 \\
 & 2 & 3.18 & 99.27 & 89.78 & 90.77 & 85.82 & 92.97 & 91.72 & 61.35 & 67.01 & 67.98 & 79.86 & 88.73 \\
 & 3 & 4.70 & 99.43 & 92.98 & 95.50 & 90.33 & 94.50 & 94.55 & 66.36 & 71.11 & 82.92 & 90.04 & 94.66 \\
 & 4 & 6.22 & 99.48 & 94.93 & 97.00 & 91.52 & 95.23 & 95.63 & 68.54 & 73.20 & 88.36 & 93.37 & 96.18 \\
 & 5 & 7.74 & 99.50 & 96.12 & 97.80 & 92.93 & 95.48 & 96.37 & 71.49 & 75.43 & 91.44 & 94.95 & 96.90 \\
 & 6 & 9.26 & 99.68 & 96.50 & 98.10 & 93.42 & 95.82 & 96.70 & 71.83 & 75.62 & 92.59 & 95.52 & 97.32 \\
 & 7 & 10.78 & 99.70 & 96.90 & 98.34 & 93.42 & 96.00 & 96.87 & 72.37 & 75.97 & 93.30 & 95.84 & 97.55 \\
 & 8 & 12.30 & 99.63 & 96.85 & 98.59 & 93.95 & 96.05 & 97.01 & 73.12 & 76.34 & 93.62 & 96.04 & 97.65 \\
 & 9 & 13.82 & 99.58 & 96.95 & 98.74 & 94.02 & 96.07 & 97.07 & 73.26 & 76.39 & 93.72 & 96.17 & 97.77 \\
 & 10 & 15.34 & 99.67 & 97.05 & 98.74 & 94.08 & 95.95 & 97.10 & 73.42 & 76.48 & 93.79 & 96.25 & 97.85 \\
 & 11 & 16.86 & 99.70 & 97.02 & 98.74 & 93.87 & 96.02 & 97.07 & 73.63 & 76.53 & 93.76 & 96.28 & 97.85 \\
 & 12 & 18.37 & 99.72 & 96.90 & 98.69 & 94.15 & 95.85 & 97.06 & 73.50 & 76.74 & 93.69 & 96.20 & 97.80 \\
\hline
\end{tabular}
\end{table*}

\begin{table*}[t]
\centering
\caption{Full per-exit results for SD-ViT with $\omega{=}0.9$ (ViT-S/12, WF4M, AdaFace).}
\label{tab:supp_w09}
\scriptsize
\setlength{\tabcolsep}{2.3pt}
\begin{tabular}{l c r|cccccc|cc|ccc}
\hline
Method & Exit & GFLOP & \multicolumn{6}{c|}{High-quality verification (\%)} & \multicolumn{2}{c|}{TinyFace} & \multicolumn{3}{c}{IJB-C (TAR@FAR)} \\
& $k$ & & LFW & AgeDB & CFP-FP & CPLFW & CALFW & Avg & R-1 & R-5 & $10^{-5}$ & $10^{-4}$ & $10^{-3}$ \\
\hline
\multirow{12}{*}{$\omega{=}0.9$} & 1 & 1.66 & 91.80 & 66.75 & 68.04 & 67.63 & 79.57 & 74.76 & 43.67 & 50.89 & 14.70 & 25.26 & 38.95 \\
 & 2 & 3.18 & 97.67 & 81.77 & 80.57 & 79.70 & 89.37 & 85.82 & 53.67 & 60.14 & 42.19 & 55.29 & 68.44 \\
 & 3 & 4.70 & 98.92 & 87.92 & 89.53 & 85.88 & 91.78 & 90.81 & 60.92 & 66.66 & 68.22 & 79.11 & 88.09 \\
 & 4 & 6.22 & 99.32 & 91.47 & 93.24 & 88.73 & 93.38 & 93.23 & 65.29 & 69.90 & 77.92 & 86.78 & 92.63 \\
 & 5 & 7.74 & 99.33 & 93.43 & 95.17 & 90.23 & 94.37 & 94.51 & 67.41 & 72.32 & 84.52 & 91.02 & 94.97 \\
 & 6 & 9.26 & 99.62 & 94.88 & 96.71 & 91.67 & 94.78 & 95.53 & 69.34 & 73.52 & 88.41 & 93.35 & 96.09 \\
 & 7 & 10.78 & 99.60 & 95.38 & 97.61 & 92.50 & 95.43 & 96.10 & 70.95 & 74.81 & 91.28 & 94.86 & 96.89 \\
 & 8 & 12.30 & 99.63 & 96.47 & 98.04 & 93.37 & 95.70 & 96.64 & 71.86 & 75.75 & 92.52 & 95.51 & 97.30 \\
 & 9 & 13.82 & 99.70 & 96.82 & 98.43 & 93.97 & 95.85 & 96.95 & 72.53 & 76.29 & 93.25 & 96.17 & 97.58 \\
 & 10 & 15.34 & 99.67 & 97.18 & 98.69 & 94.02 & 96.00 & 97.11 & 73.20 & 76.50 & 93.85 & 96.34 & 97.72 \\
 & 11 & 16.86 & 99.77 & 97.22 & 98.86 & 94.12 & 95.97 & 97.19 & 73.28 & 76.69 & 94.29 & 96.62 & 97.91 \\
 & 12 & 18.37 & 99.77 & 97.33 & 98.96 & 94.50 & 96.03 & 97.32 & 73.77 & 77.12 & 94.20 & 96.73 & 97.93 \\
\hline
\end{tabular}
\end{table*}

\begin{table*}[t]
\centering
\caption{Full per-exit results for \textbf{ViT-S/15} on WebFace4M (AdaFace).}
\label{tab:supp_d15}
\scriptsize
\setlength{\tabcolsep}{2.3pt}
\begin{tabular}{l c r|cccccc|cc|ccc}
\hline
Method & Exit & GFLOP & \multicolumn{6}{c|}{High-quality verification (\%)} & \multicolumn{2}{c|}{TinyFace} & \multicolumn{3}{c}{IJB-C (TAR@FAR)} \\
& $k$ & & LFW & AgeDB & CFP-FP & CPLFW & CALFW & Avg & R-1 & R-5 & $10^{-5}$ & $10^{-4}$ & $10^{-3}$ \\
\hline
\multirow{15}{*}{Baseline} & 1 & 1.66 & 70.87 & 53.12 & 62.46 & 56.03 & 56.78 & 59.85 & 30.02 & 38.41 & 2.38 & 5.78 & 13.08 \\
 & 2 & 3.18 & 77.20 & 56.43 & 62.21 & 56.13 & 62.22 & 62.84 & 40.72 & 49.92 & 4.21 & 9.77 & 20.14 \\
 & 3 & 4.70 & 83.28 & 60.70 & 63.96 & 58.75 & 66.72 & 66.68 & 45.52 & 54.56 & 7.99 & 16.32 & 29.85 \\
 & 4 & 6.22 & 89.38 & 67.05 & 70.97 & 67.55 & 71.87 & 73.36 & 51.29 & 60.03 & 17.36 & 29.96 & 46.22 \\
 & 5 & 7.74 & 93.63 & 71.05 & 77.09 & 73.18 & 78.70 & 78.73 & 56.12 & 64.30 & 27.66 & 41.65 & 58.16 \\
 & 6 & 9.26 & 96.78 & 76.63 & 86.13 & 80.82 & 83.72 & 84.82 & 59.95 & 67.09 & 45.90 & 59.95 & 74.61 \\
\hline
\end{tabular}
\end{table*}

\begin{table*}[t]
\centering
\caption{Full per-exit results for SD-ViT supervising \emph{fixed} exits $\{3,6\}$ (ViT-S/12, WF4M, AdaFace).}
\label{tab:supp_e36}
\scriptsize
\setlength{\tabcolsep}{2.3pt}
\begin{tabular}{l c r|cccccc|cc|ccc}
\hline
Method & Exit & GFLOP & \multicolumn{6}{c|}{High-quality verification (\%)} & \multicolumn{2}{c|}{TinyFace} & \multicolumn{3}{c}{IJB-C (TAR@FAR)} \\
& $k$ & & LFW & AgeDB & CFP-FP & CPLFW & CALFW & Avg & R-1 & R-5 & $10^{-5}$ & $10^{-4}$ & $10^{-3}$ \\
\hline
\multirow{12}{*}{exits\,$\{3,6\}$} & 1 & 1.66 & 79.68 & 58.25 & 63.24 & 59.77 & 63.12 & 64.81 & 38.09 & 46.94 & 7.17 & 14.70 & 27.33 \\
 & 2 & 3.18 & 94.98 & 75.92 & 76.51 & 74.13 & 84.37 & 81.18 & 53.00 & 59.90 & 34.21 & 48.65 & 63.53 \\
 & 3 & 4.70 & 99.57 & 94.22 & 96.11 & 91.50 & 94.97 & 95.27 & 67.06 & 71.94 & 85.86 & 91.89 & 95.53 \\
 & 4 & 6.22 & 99.52 & 93.25 & 95.37 & 91.08 & 94.20 & 94.68 & 66.74 & 71.16 & 83.28 & 90.71 & 94.90 \\
 & 5 & 7.74 & 99.60 & 94.72 & 96.76 & 91.85 & 94.73 & 95.53 & 67.97 & 72.51 & 87.35 & 92.73 & 95.81 \\
 & 6 & 9.26 & 99.58 & 96.12 & 98.26 & 93.33 & 95.58 & 96.57 & 71.75 & 75.64 & 92.44 & 95.57 & 97.38 \\
 & 7 & 10.78 & 99.60 & 96.17 & 98.04 & 92.98 & 95.58 & 96.47 & 71.35 & 74.87 & 91.97 & 95.46 & 97.27 \\
 & 8 & 12.30 & 99.65 & 95.92 & 97.96 & 93.33 & 95.42 & 96.46 & 71.54 & 74.95 & 91.13 & 94.95 & 97.04 \\
 & 9 & 13.82 & 99.70 & 96.33 & 98.19 & 93.73 & 95.63 & 96.72 & 72.21 & 75.59 & 92.06 & 95.41 & 97.31 \\
 & 10 & 15.34 & 99.68 & 96.77 & 98.23 & 93.83 & 95.80 & 96.86 & 72.48 & 76.07 & 92.64 & 95.92 & 97.60 \\
 & 11 & 16.86 & 99.73 & 97.05 & 98.76 & 94.03 & 95.87 & 97.09 & 72.99 & 76.66 & 93.79 & 96.32 & 97.84 \\
 & 12 & 18.37 & 99.83 & 97.08 & 98.77 & 94.35 & 95.98 & 97.20 & 73.39 & 77.04 & 93.85 & 96.36 & 97.83 \\
\hline
\end{tabular}
\end{table*}

\begin{table*}[t]
\centering
\caption{Full per-exit results for SD-ViT supervising the \emph{fixed} exit $\{6\}$ only (ViT-S/12, WF4M, AdaFace).}
\label{tab:supp_e6}
\scriptsize
\setlength{\tabcolsep}{2.3pt}
\begin{tabular}{l c r|cccccc|cc|ccc}
\hline
Method & Exit & GFLOP & \multicolumn{6}{c|}{High-quality verification (\%)} & \multicolumn{2}{c|}{TinyFace} & \multicolumn{3}{c}{IJB-C (TAR@FAR)} \\
& $k$ & & LFW & AgeDB & CFP-FP & CPLFW & CALFW & Avg & R-1 & R-5 & $10^{-5}$ & $10^{-4}$ & $10^{-3}$ \\
\hline
\multirow{12}{*}{exit\,$\{6\}$} & 1 & 1.66 & 74.62 & 55.22 & 62.21 & 56.73 & 59.68 & 61.69 & 31.81 & 40.77 & 5.94 & 13.26 & 25.07 \\
 & 2 & 3.18 & 84.63 & 61.70 & 64.70 & 59.95 & 67.57 & 67.71 & 43.70 & 52.07 & 9.41 & 17.88 & 31.25 \\
 & 3 & 4.70 & 91.92 & 70.22 & 73.09 & 68.62 & 77.73 & 76.32 & 51.29 & 59.36 & 25.07 & 38.13 & 53.41 \\
 & 4 & 6.22 & 97.67 & 84.00 & 86.49 & 82.20 & 89.15 & 87.90 & 59.90 & 66.42 & 58.26 & 71.38 & 82.54 \\
 & 5 & 7.74 & 99.52 & 92.82 & 95.43 & 90.80 & 94.25 & 94.56 & 66.55 & 71.24 & 84.26 & 90.55 & 94.75 \\
 & 6 & 9.26 & 99.72 & 96.92 & 98.59 & 93.88 & 95.93 & 97.01 & 72.29 & 75.35 & 93.72 & 96.07 & 97.59 \\
 & 7 & 10.78 & 99.70 & 96.78 & 98.53 & 94.07 & 95.95 & 97.01 & 71.92 & 75.48 & 93.39 & 95.97 & 97.57 \\
 & 8 & 12.30 & 99.67 & 96.98 & 98.49 & 93.77 & 95.90 & 96.96 & 71.86 & 75.51 & 93.52 & 96.06 & 97.60 \\
 & 9 & 13.82 & 99.70 & 96.90 & 98.67 & 93.88 & 95.93 & 97.02 & 72.00 & 75.48 & 93.68 & 96.18 & 97.66 \\
 & 10 & 15.34 & 99.67 & 97.07 & 98.76 & 94.37 & 96.02 & 97.18 & 72.77 & 75.99 & 93.69 & 96.31 & 97.77 \\
 & 11 & 16.86 & 99.77 & 97.08 & 98.90 & 94.17 & 96.15 & 97.21 & 73.07 & 76.42 & 94.20 & 96.57 & 97.89 \\
 & 12 & 18.37 & 99.77 & 97.32 & 98.93 & 94.42 & 96.12 & 97.31 & 72.83 & 76.31 & 94.17 & 96.63 & 97.92 \\
\hline
\end{tabular}
\end{table*}

\begin{table*}[t]
\centering
\caption{Full per-exit results for \textbf{knowledge distillation with an external teacher} (ViT-S/12, WF4M, AdaFace): a separately trained full-depth teacher, used either as an MSE target for the final embedding or by freezing the class prototypes $W$ at its full-depth solution.}
\label{tab:supp_kd}
\scriptsize
\setlength{\tabcolsep}{2.3pt}
\begin{tabular}{l c r|cccccc|cc|ccc}
\hline
Method & Exit & GFLOP & \multicolumn{6}{c|}{High-quality verification (\%)} & \multicolumn{2}{c|}{TinyFace} & \multicolumn{3}{c}{IJB-C (TAR@FAR)} \\
& $k$ & & LFW & AgeDB & CFP-FP & CPLFW & CALFW & Avg & R-1 & R-5 & $10^{-5}$ & $10^{-4}$ & $10^{-3}$ \\
\hline
\multirow{12}{*}{Embed. MSE} & 1 & 1.66 & 72.77 & 53.85 & 63.41 & 56.33 & 57.03 & 60.68 & 30.23 & 38.33 & 2.85 & 6.65 & 14.81 \\
 & 2 & 3.18 & 79.68 & 58.30 & 62.07 & 56.95 & 63.03 & 64.01 & 39.46 & 48.82 & 5.49 & 11.95 & 23.47 \\
 & 3 & 4.70 & 84.78 & 62.32 & 64.93 & 61.92 & 68.68 & 68.53 & 46.59 & 55.58 & 12.02 & 21.48 & 35.94 \\
 & 4 & 6.22 & 89.40 & 66.23 & 72.54 & 67.88 & 74.10 & 74.03 & 53.73 & 61.40 & 22.42 & 35.07 & 49.53 \\
 & 5 & 7.74 & 93.37 & 71.32 & 80.49 & 75.15 & 79.52 & 79.97 & 58.07 & 65.69 & 34.98 & 49.43 & 63.82 \\
 & 6 & 9.26 & 97.42 & 78.78 & 90.13 & 84.80 & 85.13 & 87.25 & 60.03 & 67.57 & 55.63 & 68.11 & 80.50 \\
\hline
\end{tabular}
\end{table*}

\begin{table*}[t]
\centering
\caption{Full per-exit results for \textbf{feature-matching objectives} (ViT-S/12, WF4M, AdaFace): SD-ViT's secondary classification loss replaced by an MSE between the intermediate and final embeddings, taken before or after the projection head.}
\label{tab:supp_selfmse}
\scriptsize
\setlength{\tabcolsep}{2.3pt}
\begin{tabular}{l c r|cccccc|cc|ccc}
\hline
Method & Exit & GFLOP & \multicolumn{6}{c|}{High-quality verification (\%)} & \multicolumn{2}{c|}{TinyFace} & \multicolumn{3}{c}{IJB-C (TAR@FAR)} \\
& $k$ & & LFW & AgeDB & CFP-FP & CPLFW & CALFW & Avg & R-1 & R-5 & $10^{-5}$ & $10^{-4}$ & $10^{-3}$ \\
\hline
\multirow{12}{*}{MSE pre-proj.} & 1 & 1.66 & 72.72 & 52.32 & 61.99 & 56.08 & 58.37 & 60.30 & 31.60 & 38.84 & 2.74 & 6.57 & 14.65 \\
 & 2 & 3.18 & 80.03 & 58.33 & 63.00 & 58.00 & 64.08 & 64.69 & 42.11 & 50.21 & 6.24 & 12.75 & 24.44 \\
 & 3 & 4.70 & 84.88 & 62.15 & 65.27 & 59.40 & 69.43 & 68.23 & 47.32 & 55.74 & 13.02 & 22.59 & 36.48 \\
 & 4 & 6.22 & 89.97 & 67.48 & 72.57 & 63.43 & 74.72 & 73.63 & 53.00 & 59.60 & 20.41 & 32.19 & 47.45 \\
 & 5 & 7.74 & 94.13 & 72.20 & 81.43 & 73.62 & 80.48 & 80.37 & 57.22 & 64.54 & 35.83 & 50.77 & 65.83 \\
 & 6 & 9.26 & 97.05 & 77.78 & 87.94 & 81.40 & 85.83 & 86.00 & 61.94 & 67.57 & 51.54 & 66.09 & 79.29 \\
\hline
\end{tabular}
\end{table*}

\begin{table*}[t]
\centering
\caption{Full per-exit results for the two \textbf{external-teacher variants that act on the sampled exit} (ViT-S/12, WF4M, AdaFace). \emph{SD, frozen $W$}: SD-ViT with the class prototypes $W$ frozen at the teacher's full-depth solution. \emph{Feature-MSE (teacher)}: margin loss on the final exit only, plus an MSE from the sampled intermediate exit to the frozen teacher's embedding, so the sampled exit receives no label loss.}
\label{tab:supp_sdteacher}
\scriptsize
\setlength{\tabcolsep}{2.3pt}
\begin{tabular}{l c r|cccccc|cc|ccc}
\hline
Method & Exit & GFLOP & \multicolumn{6}{c|}{High-quality verification (\%)} & \multicolumn{2}{c|}{TinyFace} & \multicolumn{3}{c}{IJB-C (TAR@FAR)} \\
& $k$ & & LFW & AgeDB & CFP-FP & CPLFW & CALFW & Avg & R-1 & R-5 & $10^{-5}$ & $10^{-4}$ & $10^{-3}$ \\
\hline
\multirow{12}{*}{SD, frozen $W$} & 1 & 1.66 & 96.15 & 77.02 & 69.80 & 71.45 & 86.40 & 80.16 & 45.92 & 51.61 & 27.69 & 40.23 & 53.89 \\
 & 2 & 3.18 & 98.95 & 89.48 & 88.00 & 84.32 & 92.98 & 90.75 & 56.63 & 61.80 & 62.89 & 73.42 & 83.46 \\
 & 3 & 4.70 & 99.38 & 93.50 & 93.90 & 89.70 & 94.42 & 94.18 & 62.98 & 67.01 & 83.39 & 89.93 & 94.21 \\
 & 4 & 6.22 & 99.58 & 95.18 & 96.24 & 91.32 & 95.03 & 95.47 & 66.50 & 70.17 & 88.74 & 93.36 & 96.04 \\
 & 5 & 7.74 & 99.63 & 96.40 & 97.94 & 92.97 & 95.62 & 96.51 & 69.10 & 73.07 & 92.02 & 95.18 & 96.98 \\
 & 6 & 9.26 & 99.63 & 97.02 & 98.40 & 93.62 & 95.70 & 96.87 & 70.98 & 74.62 & 93.26 & 95.93 & 97.41 \\
\hline
\end{tabular}
\end{table*}

\begin{table}[t]
\centering
\caption{Gallery-fixed verification on the five high-quality sets, full per-exit
(ViT-S/12, WF4M). The probe embedding is taken from exit $k$ and matched against a
gallery embedding from the final block. This is the deployment setting behind
Sec.~4.4 of the main paper: a single full-depth gallery, queried by early-exit
probes. TinyFace and IJB-C under the same protocol are in Table~\ref{tab:gf_hard}.}
\label{tab:gf_full}
\scriptsize
\setlength{\tabcolsep}{2.3pt}
\begin{tabular}{l c r|cccccc}
\hline
Method & Exit & GFLOP & LFW & AgeDB & CFP-FP & CPLFW & CALFW & Avg \\
\hline
\multirow{12}{*}{Baseline} & 1 & 1.66 & 66.22 & 60.55 & 58.53 & 59.85 & 60.20 & 61.07 \\
 & 2 & 3.18 & 70.05 & 60.72 & 60.93 & 63.10 & 63.45 & 63.65 \\
 & 3 & 4.70 & 74.57 & 63.17 & 64.76 & 65.72 & 65.68 & 66.78 \\
 & 4 & 6.22 & 82.08 & 69.10 & 71.80 & 69.83 & 72.57 & 73.08 \\
 & 5 & 7.74 & 90.23 & 76.57 & 77.97 & 77.97 & 81.28 & 80.80 \\
 & 6 & 9.26 & 96.25 & 84.92 & 87.63 & 85.80 & 88.22 & 88.56 \\
 & 7 & 10.78 & 98.43 & 90.12 & 93.31 & 90.88 & 92.10 & 92.97 \\
\hline
\end{tabular}
\end{table}

\begin{table}[t]
\centering
\caption{Gallery-fixed evaluation on TinyFace and IJB-C, full per-exit (ViT-S/12, WF4M), companion to Table~\ref{tab:gf_full}. The probe is taken from exit $k$ and matched against a final-block gallery. A value of $0.00$ is a genuine collapse, not a missing entry: against a fixed full-depth gallery the baseline is unusable on these harder benchmarks until late depths, whereas SD-ViT reaches $59.31$ TinyFace Rank-1 and $93.06$ IJB-C TAR at exit~3.}
\label{tab:gf_hard}
\scriptsize
\setlength{\tabcolsep}{3.5pt}
\begin{tabular}{l c r|cc|ccc}
\hline
Method & Exit & GFLOP & \multicolumn{2}{c|}{TinyFace} & \multicolumn{3}{c}{IJB-C (TAR@FAR)} \\
& $k$ & & R-1 & R-5 & $10^{-5}$ & $10^{-4}$ & $10^{-3}$ \\
\hline
\multirow{12}{*}{Baseline} & 1 & 1.66 & 0.00 & 0.00 & 0.02 & 0.05 & 0.53 \\
 & 2 & 3.18 & 0.00 & 0.00 & 0.01 & 0.04 & 0.95 \\
 & 3 & 4.70 & 0.00 & 0.00 & 0.01 & 0.02 & 1.57 \\
 & 4 & 6.22 & 0.00 & 0.00 & 0.05 & 0.69 & 4.94 \\
 & 5 & 7.74 & 0.00 & 0.08 & 2.72 & 7.43 & 18.34 \\
 & 6 & 9.26 & 0.75 & 1.39 & 19.07 & 34.63 & 54.38 \\
\hline
\end{tabular}
\end{table}

\begin{table}[t]
\centering
\caption{Per-dataset cosine similarity between each exit's embedding and the final-block embedding, for \textbf{Baseline} (ViT-S/12, WF4M). Averaged over each benchmark's images. Exit 12 is omitted (similarity to itself is $1$). These per-dataset values are summarised by their average in Fig.~2(h) of the main paper.}
\label{tab:sim_ViT-S}
\footnotesize
\setlength{\tabcolsep}{2.7pt}
\begin{tabular}{c|ccccccc}
\hline
Exit $k$ & LFW & AgeDB & CFP-FP & CPLFW & CALFW & TinyFace & IJB-C \\
\hline
1 & 0.074 & 0.082 & 0.049 & 0.068 & 0.072 & 0.169 & 0.056 \\
2 & 0.095 & 0.100 & 0.068 & 0.087 & 0.092 & 0.215 & 0.084 \\
3 & 0.122 & 0.128 & 0.092 & 0.112 & 0.120 & 0.260 & 0.124 \\
4 & 0.178 & 0.186 & 0.138 & 0.159 & 0.175 & 0.325 & 0.185 \\
5 & 0.274 & 0.274 & 0.214 & 0.239 & 0.270 & 0.403 & 0.293 \\
6 & 0.410 & 0.398 & 0.329 & 0.359 & 0.405 & 0.501 & 0.412 \\
7 & 0.549 & 0.528 & 0.461 & 0.494 & 0.544 & 0.614 & 0.534 \\
8 & 0.666 & 0.642 & 0.583 & 0.614 & 0.663 & 0.705 & 0.645 \\
9 & 0.800 & 0.775 & 0.743 & 0.768 & 0.799 & 0.828 & 0.779 \\
10 & 0.924 & 0.908 & 0.894 & 0.902 & 0.925 & 0.910 & 0.907 \\
11 & 0.990 & 0.986 & 0.984 & 0.983 & 0.991 & 0.974 & 0.984 \\
\hline
\end{tabular}
\end{table}

\begin{table}[t]
\centering
\caption{Per-dataset cosine similarity between each exit's embedding and the final-block embedding, for \textbf{Feature-MSE} (MSE post-proj., ViT-S/12, WF4M). Averaged over each benchmark's images. Exit 12 is omitted (similarity to itself is $1$). These per-dataset values are summarised by their average in Fig.~2(h) of the main paper.}
\label{tab:sim_ViT-SSDMSE}
\footnotesize
\setlength{\tabcolsep}{2.7pt}
\begin{tabular}{c|ccccccc}
\hline
Exit $k$ & LFW & AgeDB & CFP-FP & CPLFW & CALFW & TinyFace & IJB-C \\
\hline
1 & 0.226 & 0.195 & 0.145 & 0.162 & 0.217 & 0.140 & 0.196 \\
2 & 0.389 & 0.340 & 0.277 & 0.291 & 0.381 & 0.251 & 0.343 \\
3 & 0.477 & 0.433 & 0.355 & 0.371 & 0.472 & 0.328 & 0.419 \\
4 & 0.565 & 0.526 & 0.453 & 0.467 & 0.563 & 0.430 & 0.504 \\
5 & 0.638 & 0.601 & 0.530 & 0.542 & 0.638 & 0.493 & 0.574 \\
6 & 0.707 & 0.675 & 0.602 & 0.613 & 0.709 & 0.552 & 0.641 \\
7 & 0.759 & 0.735 & 0.672 & 0.680 & 0.762 & 0.632 & 0.705 \\
8 & 0.825 & 0.799 & 0.767 & 0.772 & 0.827 & 0.723 & 0.787 \\
9 & 0.888 & 0.865 & 0.850 & 0.853 & 0.889 & 0.799 & 0.861 \\
10 & 0.942 & 0.918 & 0.919 & 0.921 & 0.941 & 0.887 & 0.926 \\
11 & 0.989 & 0.984 & 0.984 & 0.983 & 0.989 & 0.963 & 0.984 \\
\hline
\end{tabular}
\end{table}

\begin{table}[t]
\centering
\caption{Per-dataset cosine similarity between each exit's embedding and the final-block embedding, for \textbf{SD-ViT (ours)} (ViT-S/12, WF4M). Averaged over each benchmark's images. Exit 12 is omitted (similarity to itself is $1$). These per-dataset values are summarised by their average in Fig.~2(h) of the main paper.}
\label{tab:sim_ViT-SSD07}
\footnotesize
\setlength{\tabcolsep}{2.7pt}
\begin{tabular}{c|ccccccc}
\hline
Exit $k$ & LFW & AgeDB & CFP-FP & CPLFW & CALFW & TinyFace & IJB-C \\
\hline
1 & 0.490 & 0.431 & 0.358 & 0.369 & 0.479 & 0.440 & 0.406 \\
2 & 0.722 & 0.679 & 0.595 & 0.597 & 0.721 & 0.580 & 0.635 \\
3 & 0.822 & 0.787 & 0.725 & 0.723 & 0.823 & 0.671 & 0.752 \\
4 & 0.881 & 0.854 & 0.813 & 0.809 & 0.882 & 0.747 & 0.827 \\
5 & 0.915 & 0.893 & 0.868 & 0.862 & 0.916 & 0.786 & 0.873 \\
6 & 0.939 & 0.919 & 0.905 & 0.901 & 0.940 & 0.842 & 0.907 \\
7 & 0.959 & 0.943 & 0.938 & 0.936 & 0.959 & 0.887 & 0.939 \\
8 & 0.972 & 0.959 & 0.957 & 0.955 & 0.972 & 0.926 & 0.957 \\
9 & 0.982 & 0.973 & 0.972 & 0.971 & 0.982 & 0.950 & 0.973 \\
10 & 0.991 & 0.985 & 0.985 & 0.984 & 0.991 & 0.969 & 0.985 \\
11 & 0.997 & 0.994 & 0.994 & 0.993 & 0.997 & 0.987 & 0.994 \\
\hline
\end{tabular}
\end{table}

\begin{table*}[t]
\centering
\caption{ViT-FREE$_{FT}$ under \emph{both} protocols (ViT-S/12, WF4M): one projection head and classifier per exit, trained on the frozen Baseline backbone with the Baseline's data and loss. \emph{self}: each exit against itself, the setting the method was designed for. \emph{gallery-fixed}: probe re-encoded at exit $k$ against a gallery enrolled once by the original full-depth model, the requirement identified in this paper, which ViT-FREE$_{FT}$ does not claim to address. Compare with SD-ViT in Tables~\ref{tab:supp_s12} and \ref{tab:gf_full}-\ref{tab:gf_hard}.}
\label{tab:supp_vitfree}
\scriptsize
\setlength{\tabcolsep}{2.3pt}
\begin{tabular}{l c r|cccccc|cc|ccc}
\hline
Protocol & Exit & GFLOP & \multicolumn{6}{c|}{High-quality verification (\%)} & \multicolumn{2}{c|}{TinyFace} & \multicolumn{3}{c}{IJB-C (TAR@FAR)} \\
& $k$ & & LFW & AgeDB & CFP-FP & CPLFW & CALFW & Avg & R-1 & R-5 & $10^{-5}$ & $10^{-4}$ & $10^{-3}$ \\
\hline
\multirow{5}{*}{self} & 1 & 1.66 & 95.87 & 78.20 & 76.83 & 72.98 & 85.65 & 81.91 & 46.19 & 51.93 & 26.66 & 41.59 & 59.93 \\
 & 3 & 4.70 & 98.42 & 85.22 & 87.59 & 82.85 & 90.48 & 88.91 & 57.56 & 63.52 & 57.73 & 71.36 & 83.27 \\
 & 6 & 9.26 & 99.58 & 93.43 & 95.84 & 90.77 & 93.77 & 94.68 & 67.09 & 71.97 & 82.22 & 89.74 & 94.69 \\
 & 9 & 13.82 & 99.77 & 96.78 & 98.60 & 93.95 & 95.67 & 96.95 & 73.20 & 76.37 & 93.51 & 96.03 & 97.53 \\
 & 12 & 18.37 & 99.78 & 97.57 & 99.00 & 94.67 & 95.93 & 97.39 & 73.39 & 76.66 & 94.52 & 96.74 & 97.96 \\
\hline
\hline
\multirow{5}{*}{gallery-fixed} & 1 & 1.66 & 94.50 & 82.93 & 88.70 & 81.88 & 86.85 & 86.97 & 8.99 & 14.57 & 37.03 & 47.68 & 60.66 \\
 & 3 & 4.70 & 96.57 & 85.38 & 90.97 & 85.48 & 89.33 & 89.55 & 16.93 & 25.13 & 50.17 & 61.78 & 73.63 \\
 & 6 & 9.26 & 99.28 & 93.42 & 96.69 & 92.05 & 94.03 & 95.09 & 52.36 & 59.17 & 83.58 & 89.09 & 93.09 \\
 & 9 & 13.82 & 99.65 & 96.53 & 98.60 & 94.13 & 95.83 & 96.95 & 70.04 & 73.69 & 93.32 & 95.75 & 97.23 \\
 & 12 & 18.37 & 99.78 & 97.13 & 98.80 & 94.12 & 95.95 & 97.16 & 72.96 & 75.89 & 94.41 & 96.51 & 97.78 \\
\hline
\end{tabular}
\end{table*}

\begin{table}[t]
\centering
\caption{Matched-compute references: ViT-S backbones of depth 3 and 6 trained from scratch with the identical setting, i.e.\ dedicated single-budget networks, against the SD-ViT exits of the same cost. Each truncated model has a single exit, its own full depth. The dedicated networks are better at their own budget, the multi-exit model covers all budgets at once, against one gallery.}
\label{tab:supp_trunc}
\footnotesize
\setlength{\tabcolsep}{3.5pt}
\begin{tabular}{l|r|ccc}
\hline
Model & GFLOP & HQ\,(5) & TinyFace R-1 & IJB-C $10^{-4}$ \\
\hline
ViT-S/3, from scratch & 4.70 & \textbf{95.86} & \textbf{69.50} & \textbf{93.78} \\
SD-ViT, exit 3        & 4.70 & 93.68 & 64.62 & 87.01 \\
\hline
ViT-S/6, from scratch & 9.26 & \textbf{96.96} & \textbf{71.83} & \textbf{95.94} \\
SD-ViT, exit 6        & 9.26 & 96.42 & 71.22 & 95.11 \\
\hline
ViT-S/12 (full)       & 18.37 & \textbf{97.43} & \textbf{74.25} & \textbf{96.89} \\
SD-ViT, exit 12       & 18.37 & 97.29 & 73.85 & 96.51 \\
\hline
\end{tabular}
\end{table}

\begin{table*}[t]
\centering
\caption{Deterministic supervision of all 11 intermediate exits (ViT-S/12, WF4M), upper block self-evaluation and lower block gallery-fixed. Compare with stochastic SD-ViT in Tables~\ref{tab:supp_s12} and \ref{tab:gf_full}-\ref{tab:gf_hard}: from exit~3 onward the two are within about a point, a gap one run per configuration cannot resolve, while at exits~1 and 2 sampling is ahead ($40.32$ vs.\ $35.69$ and $74.37$ vs.\ $71.37$ IJB-C, self-evaluation). All-exit supervision costs $88.3$ A100 GPU-hours against $68.6$, i.e.\ $7.4\times$ the auxiliary cost over the $65.5$ GPU-h baseline.}
\label{tab:supp_allexit}
\scriptsize
\setlength{\tabcolsep}{2.3pt}
\begin{tabular}{l c r|cccccc|cc|ccc}
\hline
Variant & Exit & GFLOP & \multicolumn{6}{c|}{High-quality verification (\%)} & \multicolumn{2}{c|}{TinyFace} & \multicolumn{3}{c}{IJB-C (TAR@FAR)} \\
& $k$ & & LFW & AgeDB & CFP-FP & CPLFW & CALFW & Avg & R-1 & R-5 & $10^{-5}$ & $10^{-4}$ & $10^{-3}$ \\
\hline
\multirow{12}{*}{all exits, self} & 1 & 1.66 & 95.40 & 72.22 & 73.20 & 72.23 & 83.68 & 79.35 & 50.48 & 57.19 & 23.51 & 35.69 & 49.47 \\
 & 2 & 3.18 & 99.08 & 87.78 & 88.64 & 84.20 & 92.00 & 90.34 & 59.20 & 65.77 & 59.21 & 71.37 & 82.32 \\
 & 3 & 4.70 & 99.38 & 92.30 & 93.86 & 89.40 & 93.83 & 93.75 & 63.92 & 69.31 & 76.96 & 86.54 & 92.74 \\
 & 4 & 6.22 & 99.55 & 94.58 & 96.10 & 91.25 & 94.63 & 95.22 & 67.14 & 71.65 & 85.37 & 91.62 & 95.51 \\
 & 5 & 7.74 & 99.62 & 95.60 & 97.27 & 92.17 & 95.12 & 95.95 & 69.82 & 73.74 & 89.49 & 93.95 & 96.56 \\
 & 6 & 9.26 & 99.58 & 96.45 & 97.74 & 92.93 & 95.42 & 96.43 & 71.67 & 75.75 & 91.45 & 95.15 & 97.05 \\
 & 7 & 10.78 & 99.70 & 96.80 & 98.39 & 93.58 & 95.90 & 96.87 & 72.34 & 76.37 & 92.65 & 95.73 & 97.45 \\
 & 8 & 12.30 & 99.75 & 97.12 & 98.40 & 94.00 & 95.92 & 97.04 & 72.99 & 76.72 & 93.59 & 96.16 & 97.69 \\
 & 9 & 13.82 & 99.78 & 97.17 & 98.89 & 94.33 & 95.93 & 97.22 & 73.28 & 76.50 & 93.82 & 96.36 & 97.77 \\
 & 10 & 15.34 & 99.72 & 97.05 & 98.93 & 94.32 & 96.10 & 97.22 & 73.31 & 76.82 & 93.96 & 96.49 & 97.85 \\
 & 11 & 16.86 & 99.80 & 97.47 & 98.87 & 94.45 & 96.07 & 97.33 & 73.47 & 76.85 & 94.03 & 96.51 & 97.88 \\
 & 12 & 18.37 & 99.80 & 97.45 & 98.83 & 94.47 & 96.02 & 97.31 & 73.23 & 76.90 & 93.96 & 96.49 & 97.84 \\
\hline
\multirow{12}{*}{all exits, gf} & 1 & 1.66 & 97.47 & 84.10 & 91.56 & 85.38 & 88.85 & 89.47 & 9.92 & 13.36 & 38.87 & 53.19 & 66.83 \\
 & 2 & 3.18 & 99.40 & 92.65 & 97.29 & 91.72 & 93.92 & 94.99 & 46.57 & 52.01 & 80.74 & 87.72 & 92.86 \\
 & 3 & 4.70 & 99.62 & 94.85 & 98.13 & 92.82 & 94.88 & 96.06 & 59.07 & 63.28 & 88.06 & 92.73 & 95.74 \\
 & 4 & 6.22 & 99.65 & 96.03 & 98.30 & 93.57 & 95.53 & 96.62 & 64.70 & 68.94 & 90.77 & 94.45 & 96.63 \\
 & 5 & 7.74 & 99.78 & 96.75 & 98.41 & 93.60 & 95.80 & 96.87 & 68.32 & 72.05 & 92.31 & 95.30 & 97.17 \\
 & 6 & 9.26 & 99.77 & 97.02 & 98.64 & 93.87 & 95.85 & 97.03 & 71.06 & 75.08 & 93.07 & 95.79 & 97.47 \\
 & 7 & 10.78 & 99.82 & 97.08 & 98.69 & 94.03 & 95.85 & 97.09 & 72.24 & 75.75 & 93.58 & 96.15 & 97.64 \\
 & 8 & 12.30 & 99.78 & 97.08 & 98.86 & 94.20 & 95.93 & 97.17 & 72.53 & 76.37 & 94.01 & 96.34 & 97.76 \\
 & 9 & 13.82 & 99.80 & 97.32 & 98.80 & 94.13 & 95.92 & 97.19 & 73.02 & 76.69 & 94.08 & 96.50 & 97.78 \\
 & 10 & 15.34 & 99.80 & 97.22 & 98.90 & 94.30 & 95.97 & 97.24 & 73.18 & 76.74 & 94.13 & 96.56 & 97.83 \\
 & 11 & 16.86 & 99.80 & 97.50 & 98.93 & 94.38 & 96.10 & 97.34 & 73.23 & 76.93 & 94.10 & 96.58 & 97.88 \\
 & 12 & 18.37 & 99.80 & 97.45 & 98.83 & 94.47 & 96.02 & 97.31 & 73.23 & 76.90 & 93.96 & 96.49 & 97.84 \\
\hline
\end{tabular}
\end{table*}

\begin{table*}[t]
\centering
\caption{Separate intermediate prototypes $W'$ (ViT-S/12, WF4M): shared head, one prototype set for the final exit and one extra set shared by all sampled intermediate exits. Upper block: self-evaluation. Lower block: gallery-fixed. Compare with SD-ViT in Tables~\ref{tab:supp_s12} and \ref{tab:gf_full}-\ref{tab:gf_hard}.}
\label{tab:supp_splitproto}
\scriptsize
\setlength{\tabcolsep}{2.3pt}
\begin{tabular}{l c r|cccccc|cc|ccc}
\hline
Variant & Exit & GFLOP & \multicolumn{6}{c|}{High-quality verification (\%)} & \multicolumn{2}{c|}{TinyFace} & \multicolumn{3}{c}{IJB-C (TAR@FAR)} \\
& $k$ & & LFW & AgeDB & CFP-FP & CPLFW & CALFW & Avg & R-1 & R-5 & $10^{-5}$ & $10^{-4}$ & $10^{-3}$ \\
\hline
\multirow{12}{*}{$W'$ self} & 1 & 1.66 & 96.28 & 75.47 & 76.94 & 74.70 & 85.25 & 81.73 & 51.80 & 58.61 & 29.83 & 42.54 & 57.22 \\
 & 2 & 3.18 & 98.92 & 88.50 & 88.96 & 84.62 & 92.20 & 90.64 & 60.11 & 66.28 & 61.92 & 74.52 & 85.55 \\
 & 3 & 4.70 & 99.33 & 92.32 & 94.19 & 89.05 & 93.62 & 93.70 & 65.64 & 70.23 & 77.99 & 86.90 & 92.97 \\
 & 4 & 6.22 & 99.60 & 94.05 & 95.57 & 90.63 & 94.55 & 94.88 & 67.41 & 71.86 & 84.48 & 91.08 & 95.22 \\
 & 5 & 7.74 & 99.43 & 95.50 & 97.29 & 92.08 & 95.10 & 95.88 & 69.61 & 74.11 & 88.42 & 93.57 & 96.38 \\
 & 6 & 9.26 & 99.62 & 95.77 & 98.09 & 92.77 & 95.43 & 96.33 & 70.90 & 75.24 & 90.41 & 94.59 & 96.89 \\
 & 7 & 10.78 & 99.57 & 96.43 & 98.21 & 93.27 & 95.67 & 96.63 & 72.40 & 76.02 & 92.11 & 95.34 & 97.31 \\
 & 8 & 12.30 & 99.68 & 96.63 & 98.54 & 93.78 & 95.63 & 96.86 & 72.96 & 76.77 & 92.90 & 95.77 & 97.56 \\
 & 9 & 13.82 & 99.77 & 96.80 & 98.54 & 94.03 & 95.90 & 97.01 & 73.26 & 76.72 & 93.65 & 96.23 & 97.71 \\
 & 10 & 15.34 & 99.77 & 96.93 & 98.80 & 94.18 & 95.92 & 97.12 & 73.50 & 76.77 & 94.06 & 96.45 & 97.83 \\
 & 11 & 16.86 & 99.80 & 97.07 & 98.91 & 94.30 & 95.95 & 97.21 & 73.71 & 76.50 & 94.28 & 96.65 & 97.93 \\
 & 12 & 18.37 & 99.82 & 97.42 & 98.91 & 94.45 & 96.02 & 97.32 & 73.39 & 76.21 & 94.68 & 96.69 & 97.87 \\
\hline
\multirow{12}{*}{$W'$ gf} & 1 & 1.66 & 97.67 & 86.55 & 92.47 & 86.85 & 90.12 & 90.73 & 9.33 & 13.30 & 40.71 & 60.96 & 74.39 \\
 & 2 & 3.18 & 99.33 & 93.00 & 96.81 & 91.05 & 93.82 & 94.80 & 44.64 & 50.24 & 78.92 & 86.59 & 92.00 \\
 & 3 & 4.70 & 99.47 & 94.85 & 97.81 & 92.83 & 94.65 & 95.92 & 56.36 & 61.08 & 86.15 & 91.76 & 94.96 \\
 & 4 & 6.22 & 99.50 & 95.72 & 98.09 & 93.42 & 95.38 & 96.42 & 61.72 & 65.99 & 89.52 & 93.48 & 95.96 \\
 & 5 & 7.74 & 99.63 & 96.20 & 98.26 & 94.02 & 95.43 & 96.71 & 65.02 & 69.29 & 91.43 & 94.64 & 96.65 \\
 & 6 & 9.26 & 99.77 & 96.68 & 98.50 & 93.95 & 95.73 & 96.93 & 67.09 & 71.27 & 92.23 & 95.11 & 96.98 \\
 & 7 & 10.78 & 99.73 & 96.87 & 98.64 & 94.02 & 95.70 & 96.99 & 69.02 & 72.96 & 93.11 & 95.61 & 97.33 \\
 & 8 & 12.30 & 99.72 & 97.00 & 98.80 & 94.20 & 95.70 & 97.08 & 70.57 & 74.09 & 93.42 & 95.90 & 97.43 \\
 & 9 & 13.82 & 99.72 & 97.02 & 98.73 & 94.40 & 95.83 & 97.14 & 71.06 & 74.60 & 93.83 & 96.15 & 97.63 \\
 & 10 & 15.34 & 99.77 & 97.08 & 98.86 & 94.27 & 95.93 & 97.18 & 72.16 & 75.21 & 94.31 & 96.42 & 97.77 \\
 & 11 & 16.86 & 99.77 & 97.10 & 99.01 & 94.43 & 96.08 & 97.28 & 72.91 & 75.83 & 94.62 & 96.64 & 97.90 \\
 & 12 & 18.37 & 99.82 & 97.42 & 98.91 & 94.45 & 96.02 & 97.32 & 73.39 & 76.21 & 94.68 & 96.69 & 97.87 \\
\hline
\end{tabular}
\end{table*}

\begin{table*}[t]
\centering
\caption{Single-threshold calibration on IJB-C, full per-exit companion to Tab.~6 of the main paper (ViT-S/12, WF4M, gallery-fixed protocol, target FAR$=10^{-4}$). $\tau$ is chosen once on the development half of the \emph{full-depth} scores and applied unchanged to every exit on the held-out half. FAR is the realized false-accept rate at that fixed $\tau$, in units of $10^{-4}$. \emph{per-exit $\tau$} re-calibrates on each exit's own development scores, and \emph{loss} is the TAR difference between the two. The last two columns are the mean genuine and impostor cosine scores on the held-out half, which explain the drift: shallower probes lower the genuine mean far more than the impostor mean, so the fixed threshold becomes conservative.}
\label{tab:supp_calib}
\footnotesize
\setlength{\tabcolsep}{4pt}
\begin{tabular}{l c c|ccc|c|cc}
\hline
 & & & \multicolumn{3}{c|}{TAR (\%)} & & \multicolumn{2}{c}{mean score} \\
Method & $\tau$ & Exit $k$ & fixed $\tau$ & per-exit $\tau$ & loss & FAR ($10^{-4}$) & genuine & impostor \\
\hline
\multirow{5}{*}{Baseline} & \multirow{5}{*}{0.366} & 1 & 0.00 & 0.04 & 0.04 & 0.00 & 0.026 & -0.011 \\
 &  & 3 & 0.00 & 0.01 & 0.01 & 0.03 & 0.068 & 0.004 \\
 &  & 6 & 19.59 & 34.19 & 14.60 & 0.12 & 0.285 & 0.057 \\
 &  & 9 & 90.61 & 92.16 & 1.55 & 0.41 & 0.561 & 0.053 \\
 &  & 12 & 96.91 & 96.91 & 0.00 & 0.96 & 0.733 & 0.034 \\
\hline
\multirow{5}{*}{Feature-MSE} & \multirow{5}{*}{0.381} & 1 & 0.00 & 16.69 & 16.69 & 0.00 & 0.113 & -0.041 \\
 &  & 3 & 29.24 & 70.80 & 41.55 & 0.00 & 0.309 & -0.012 \\
 &  & 6 & 79.56 & 90.48 & 10.93 & 0.01 & 0.496 & 0.017 \\
 &  & 9 & 94.36 & 95.47 & 1.11 & 0.31 & 0.645 & 0.047 \\
 &  & 12 & 96.78 & 96.78 & 0.00 & 0.98 & 0.740 & 0.044 \\
\hline
\multirow{5}{*}{SD-ViT (ours)} & \multirow{5}{*}{0.382} & 1 & 15.66 & 63.81 & 48.15 & 0.01 & 0.269 & -0.018 \\
 &  & 3 & 89.08 & 93.05 & 3.97 & 0.13 & 0.585 & 0.016 \\
 &  & 6 & 95.15 & 95.78 & 0.64 & 0.43 & 0.688 & 0.033 \\
 &  & 9 & 96.15 & 96.34 & 0.19 & 0.75 & 0.721 & 0.039 \\
 &  & 12 & 96.48 & 96.48 & 0.00 & 1.02 & 0.731 & 0.044 \\
\hline
\end{tabular}
\end{table*}

\begin{table}[t]
\centering
\caption{Measured batch-1 forward latency and peak allocated memory per exit on a single NVIDIA GeForce RTX 4060 Ti, mean over 200 timed iterations after 50 warm-up iterations. Latency grows with the number of blocks actually executed, while peak memory is nearly flat because the shared head and the resident weights dominate it. SD-ViT and the Baseline are architecturally identical, so these figures apply to both. - marks a depth the backbone does not have.}
\label{tab:supp_lat}
\footnotesize
\setlength{\tabcolsep}{4pt}
\begin{tabular}{c|cc|cc|cc}
\hline
 & \multicolumn{2}{c|}{ViT-S/12} & \multicolumn{2}{c|}{ViT-S/15} & \multicolumn{2}{c}{ViT-B/24} \\
Exit $k$ & ms & MiB & ms & MiB & ms & MiB \\
\hline
1 & 1.04 & 385 & 1.03 & 427 & 1.00 & 454 \\
2 & 1.28 & 386 & 1.28 & 428 & 1.22 & 456 \\
3 & 1.52 & 386 & 1.52 & 428 & 1.44 & 456 \\
4 & 1.76 & 386 & 1.77 & 428 & 1.66 & 456 \\
5 & 2.00 & 386 & 2.01 & 428 & 1.86 & 456 \\
6 & 2.24 & 386 & 2.25 & 428 & 2.08 & 456 \\
7 & 2.49 & 386 & 2.50 & 428 & 2.29 & 456 \\
8 & 2.74 & 386 & 2.75 & 428 & 2.50 & 456 \\
9 & 2.99 & 386 & 3.00 & 428 & 2.72 & 456 \\
10 & 3.24 & 386 & 3.25 & 428 & 2.94 & 456 \\
11 & 3.50 & 386 & 3.50 & 428 & 3.16 & 456 \\
12 & 3.75 & 386 & 3.76 & 428 & 3.37 & 456 \\
13 & - & - & 4.01 & 428 & 3.60 & 456 \\
14 & - & - & 4.27 & 428 & 3.81 & 456 \\
15 & - & - & 4.52 & 428 & 4.03 & 456 \\
16 & - & - & - & - & 4.25 & 456 \\
17 & - & - & - & - & 4.47 & 456 \\
18 & - & - & - & - & 4.69 & 456 \\
19 & - & - & - & - & 4.91 & 456 \\
20 & - & - & - & - & 5.13 & 456 \\
21 & - & - & - & - & 5.35 & 456 \\
22 & - & - & - & - & 5.57 & 456 \\
23 & - & - & - & - & 5.79 & 456 \\
24 & - & - & - & - & 6.01 & 456 \\
\hline
\end{tabular}
\end{table}

\begin{table*}[t]
\centering
\caption{KP-RPE on WebFace4M, full per-exit results on the five high-quality sets, TinyFace Rank-1, and IJB-C TAR@FAR=$10^{-4}$. The pattern matches the plain ViT-S/12 of Table~\ref{tab:supp_s12}: at exits 3 / 6 / 12 the objective gives $94.11$ / $96.56$ / $97.37$ here against $93.68$ / $96.42$ / $97.29$ there. GFLOPs are measured: the keypoint-relative encoding projects the five keypoints, taken relative to the patch-center grid, once per image into the per-block, per-head attention biases, which adds $0.018$ GFLOPs at every exit, $+1.1\%$ at exit~1 falling to $+0.1\%$ at full depth, over the plain ViT-S/12 of Table~\ref{tab:cost}.}
\label{tab:supp_kprpe}
\scriptsize
\setlength{\tabcolsep}{3pt}
\begin{tabular}{l c r|cccccc|cc}
\hline
Method & Exit & GFLOP & LFW & AgeDB & CFP-FP & CPLFW & CALFW & Avg & TF R-1 & IJB-C $10^{-4}$ \\
\hline
\multirow{12}{*}{Baseline} & 1 & 1.68 & 70.50 & 52.07 & 61.79 & 55.90 & 57.27 & 59.50 & 29.64 & 5.52 \\
 & 2 & 3.20 & 74.73 & 53.67 & 62.89 & 55.73 & 59.82 & 61.37 & 36.19 & 7.63 \\
 & 3 & 4.72 & 81.12 & 59.33 & 62.83 & 56.82 & 63.48 & 64.72 & 42.41 & 12.67 \\
 & 4 & 6.24 & 86.03 & 63.63 & 67.37 & 60.23 & 70.28 & 69.51 & 48.15 & 21.48 \\
 & 5 & 7.76 & 91.53 & 68.17 & 76.19 & 69.70 & 76.37 & 76.39 & 53.27 & 36.30 \\
 & 6 & 9.28 & 95.62 & 74.77 & 84.84 & 77.47 & 82.12 & 82.96 & 58.91 & 53.52 \\
 & 7 & 10.80 & 97.80 & 81.63 & 89.71 & 84.77 & 87.50 & 88.28 & 61.59 & 70.86 \\
 & 8 & 12.32 & 98.85 & 88.03 & 93.61 & 89.05 & 91.12 & 92.13 & 66.17 & 82.83 \\
 & 9 & 13.84 & 99.55 & 93.08 & 96.63 & 92.60 & 94.12 & 95.20 & 69.39 & 92.33 \\
 & 10 & 15.35 & 99.72 & 96.87 & 98.49 & 94.33 & 95.78 & 97.04 & 73.23 & 95.34 \\
 & 11 & 16.87 & 99.80 & 97.60 & 98.91 & 94.90 & 96.08 & 97.46 & 75.08 & 96.91 \\
 & 12 & 18.39 & 99.83 & 97.62 & 99.11 & 94.90 & 95.85 & 97.46 & 74.84 & 97.03 \\
\hline
\multirow{12}{*}{KP-RPE $+$ SD} & 1 & 1.68 & 96.12 & 76.38 & 75.11 & 73.90 & 86.20 & 81.54 & 51.56 & 41.66 \\
 & 2 & 3.20 & 99.03 & 88.22 & 89.40 & 84.95 & 92.50 & 90.82 & 60.70 & 75.62 \\
 & 3 & 4.72 & 99.43 & 92.87 & 94.30 & 89.90 & 94.03 & 94.11 & 66.20 & 88.19 \\
 & 4 & 6.24 & 99.60 & 94.77 & 96.54 & 91.45 & 95.00 & 95.47 & 68.88 & 92.30 \\
 & 5 & 7.76 & 99.62 & 95.57 & 97.67 & 92.87 & 95.17 & 96.18 & 70.68 & 94.45 \\
 & 6 & 9.28 & 99.63 & 96.20 & 98.26 & 93.13 & 95.58 & 96.56 & 72.67 & 95.27 \\
 & 7 & 10.80 & 99.72 & 96.40 & 98.46 & 94.02 & 96.07 & 96.93 & 73.77 & 95.90 \\
 & 8 & 12.32 & 99.73 & 96.78 & 98.70 & 94.30 & 96.12 & 97.13 & 73.93 & 96.32 \\
 & 9 & 13.84 & 99.75 & 97.12 & 98.86 & 94.32 & 96.13 & 97.23 & 74.17 & 96.52 \\
 & 10 & 15.35 & 99.73 & 97.32 & 98.81 & 94.30 & 96.15 & 97.26 & 74.30 & 96.61 \\
 & 11 & 16.87 & 99.73 & 97.28 & 98.83 & 94.37 & 96.13 & 97.27 & 74.73 & 96.69 \\
 & 12 & 18.39 & 99.80 & 97.32 & 98.94 & 94.52 & 96.25 & 97.37 & 74.52 & 96.67 \\
\hline
\end{tabular}
\end{table*}

\end{document}